%% file: iclr2026_conference.tex
\documentclass{article} 
\usepackage{iclr2026_conference,times}
\usepackage{xcolor}
\usepackage{graphicx}
\usepackage{placeins} 
\usepackage{float}
\usepackage{subcaption}

\input{math_commands.tex}

\usepackage{hyperref}
\usepackage{url}
\usepackage{subcaption}
\usepackage{booktabs} 

\title{Hedgementation = Hedgerow Segmentation: \\ A Remote Sensing Benchmark}

\author{%
Nathan Senyard$^{1}$ \quad Salem Hamdani$^{3*}$ \quad Astrid Zhang$^{1*}$ \quad Derek Wang$^{1}$ \\
\vspace{2pt}
\textbf{Evan Shelhamer}$^{1,4+}$ \quad \textbf{Mathias Lécuyer}$^{1+}$ \quad \textbf{Joséphine Gantois}$^{2+}$ \\
\vspace{2pt}
$^1$UBC CS \quad $^2$UBC IRES \& FRE \quad $^3$ INSAT Tunisia \quad $^4$Vector Institute \\
$^{*}$equal contribution $^{+}$equal advising
}

\iclrfinalcopy 
\begin{document}

\maketitle

\vspace{-1em}

\begin{abstract}
We propose Hedgementation: a new benchmark to evaluate machine learning models for hedgerow mapping from remote sensing data at country scale and 10m$^2$ spatial resolution.
We combine and harmonize multiple remote sensing data products and ground truth labels sourced from a hedgerow inventory in France. 
We measure the ability of three baseline models to generalize across spatial distance, and across climatic zones, a more explicitly challenging task.
Our benchmark tests both supervised and self-supervised learning approaches for remote sensing, applied to tracking fine-scale features of high agricultural importance.
The code to reproduce the benchmark and baselines results is available at \url{https://github.com/hedgementation/hedgementation}.
\end{abstract}

\section{Introduction: Hedgerows and their Agricultural Importance}
\label{sec:intro}
\input{text/introduction}

\section{Hedgementation Benchmark}
\label{sec:hedgementation}

\subsection{The Hedgerow Dataset: Imagery, Embeddings, and Labels}
\label{sec:dataset}
\input{text/dataset}

\subsection{The Hedgerow Detection Task and its Evaluation}
\label{sec:benchmark}
\input{text/benchmark}

\section{Experiments: Baselines, Results, and Discussion}
\label{sec:results}
\input{text/results}

\input{text/discussion-short}

\section*{Acknowledgments}
We thank Perrine Porcher for discussions and experiments during her Vector internship about the PASTIS U-TAE model and its variations as applied to agricultural mapping.
Jos\'ephine Gantois acknowledges the support of the Natural Sciences and Engineering Research Council of Canada (NSERC), funding reference number ALLRP 588517 - 23.
Salem Hamdani acknowledges financial support from the Mitacs Globalink Research Internship program.
Mathias Lécuyer acknowledges financial support from the Natural Sciences and Engineering Research Council
of Canada (NSERC), reference number RGPIN-2022-04469.
Evan Shelhamer acknowledges financial support from a Canada CIFAR AI Chair and the Natural Sciences and Engineering Research Council of Canada (NSERC) Discovery Grant (RGPIN-2025-06878).
This research was enabled in part by computational support provided by the Digital Research Alliance of Canada (alliancecan.ca).
Resources used in preparing this research were provided, in part, by the Province of Ontario, the Government of Canada through CIFAR, and companies sponsoring the Vector Institute.

\bibliography{iclr2026_conference}
\bibliographystyle{iclr2026_conference}

\newpage
\appendix

\section{Data and Methods}

\subsection{Hedgerow Labels}
\label{appendix:data}
\input{text/appendix/app-data.tex}

\subsection{Pre-processing Inputs for the Fields of the World Model}
\label{appendix:ftw-arch}
\input{text/appendix/app-ftw}

\section{Related Work, Limitations, and Next Steps}
\label{appendix:discussion}
\input{text/discussion-long}

\section{Qualitative Results: Predictions}
\label{sec:results-visual}
\input{text/visual-results}

\section{Supplementary Results: Generalization over Space}
\label{appendix:space}

\subsection{PASTIS UTAE}
\label{appendix:pastis-results-space}
\input{text/appendix/results_pastis_space}

\subsection{Fields of The World}
\label{appendix:ftw}
\input{text/appendix/results_ftw_space}

\subsection{KNN + Embeddings}
\label{appendix:knn-results}
\input{text/appendix/results_knn_space}

\subsection{Random Forest + Embeddings}
\label{appendix:rf-results}
\input{text/appendix/results_rf_space}

\subsection{Logistic Regression + Embeddings}
\label{appendix:rf-results}
\input{text/appendix/results_logistic_regression_space}

\section{Supplementary Results: Generalization over Agriculturally Relevant Climatic Zones}
\label{appendix:climatic-zones}

\subsection{PASTIS UTAE}
\label{appendix:pastis-results-climatic-zones}
\input{text/appendix/results_pastis_climate-zones}

\subsection{Fields of The Wold}
\label{appendix:ftw-results-climatic-zones}
\input{text/appendix/results_ftw_climate-zones}

\section{Supplementary Results: Performance in Agricultural Geographies}
\label{appendix:ag-perf}
\input{text/appendix/ag_exps}

\subsection{PASTIS UTAE}
\label{appendix:pastis-results-ag}
\input{text/appendix/results_pastis_ag}

\subsection{Fields of The World}
\label{appendix:ftw-results-ag}
\input{text/appendix/results_ftw_ag}

\end{document}

%% file: math_commands.tex
\usepackage{amsmath,amsfonts,bm}

\def\eqref#1{equation~\ref{#1}}

\def\1{\bm{1}}

\DeclareMathAlphabet{\mathsfit}{\encodingdefault}{\sfdefault}{m}{sl}
\SetMathAlphabet{\mathsfit}{bold}{\encodingdefault}{\sfdefault}{bx}{n}



%% file: text/introduction.tex
Small patches of natural habitat like hedgerows play a critical role in agricultural landscapes, as they support a range of biodiversity outcomes, carbon storage, and ecosystem services \citep{biffi2023planting, asbjornsen2014targeting, montgomery2020hedgerows, albrecht2020effectiveness}.
It is well-known that hedgerows have seen sharp and ongoing declines, in line with shifts in agricultural systems \citep{baudry2000hedgerows, de2023haie}.
Yet, little data exists on their remaining extent and dynamics, as it is costly to collect at large scales. 
Existing inventory efforts are geographically and temporally limited (two years of inventory in France and one year in England are the most prominent).
Satellite imagery offers a promising solution to fill this data gap \citep{ha2019shelterbelt, faucqueur2019new}, especially when combined with deep learning models \citep{liu2023overlooked, muro2025hedgerow}. 
However progress is limited by the sparsity of labels, unknown or poor transferability of existing models to new regions, and lack of access and reproducibility of the existing learning methods.

We propose a new benchmark called Hedgementation, to evaluate hedgerow detection from remote sensing data across France (\S\ref{sec:hedgementation}).
Hedgementation builds a dataset of imagery, embeddings, and labels: it curates and combines harmonized imagery from the Sentinel-2 satellite~\citep{sentinel2harmonized}, embedding data from Alpha Earth Foundations~\citep{aef} or AEF, and hedgerow labels from the BD Haie inventory produced by the French IGN institute~\citep{BDhaie2024} (\S\ref{sec:dataset}).
Based on this data, we design a benchmark for the hedgerow detection task. 
We partition the data to evaluate how models generalize across space and across agriculturally-relevant climatic zones (\S\ref{sec:benchmark}).

We evaluate three baselines on Hedgementation (\S\ref{sec:results}): two deep networks, U-TAE~\citep{pastis} and FTW~\cite{ftw}), and one nearest neighbor model over the AEF embeddings.
The initial results of the baselines show promise, with the best segmentation accuracy (intersection-over-union) reaching ${>}40\%$, and better performance in agricultural areas and in temperate zones, but underline the need for more progress to achieve accurate mapping.

The code to automatically create our Hedgementation benchmark, and replicate our benchmark results, is publicly available at \url{https://github.com/hedgementation/hedgementation}.

%% file: text/dataset.tex
We combine several data sources to create a hedgerow dataset in France, where imagery and labels are publicly available.
The satellite imagery comes from the harmonized, surface reflectance, Sentinel-2 data product \citep{sentinel2harmonized}, accessed through Google Earth Engine (GEE) \citep{gorelick2017google}. Sentinel-2 provides 10-m resolution images of multispectral bands at 5-day intervals.
We use the 10 non-atmospheric bands, following prior work in self-supervised learning for remote sensing ~\citep{garnot2021panoptic,tseng2025galileo}.
We gather all images acquired Sept. 2021–Oct. 2022, and drop images with more than 20\% cloud cover.

The embeddings for Hedgementation are loaded from the AEF collection~\citep{aef} on GEE. 
These are provided at a 10-m resolution in space, and at an annual resolution in time, where each ``annualized'' representation summarizes the full data over the calendar year. 
We collect the embeddings for 2022 for the highest overlap with the temporal interval of our satellite imagery.
For any 10-m resolution pixel, we can thus load either the satellite imagery data product, with sensed spectral bands, or the embeddings data product, with learned bands.

We pair these remote sensing inputs with hedgerow labels from version 2 of the French hedgerow inventory BD Haie \citep{BDhaie2024}, which is unique in scope, accuracy, and public availability. These data are produced from aerial photo interpretation, validated by individual farmers in agricultural zones, and supplemented with automatic image segmentation in other areas \citep{Commagnac_2024_BDHaieV2}. They are representative of the status of hedgerows in 2020-2022 (Appendix \ref{appendix:data}).

The data cover all of France and contain 20,529,039 hedgerows, represented as line geometries with no width attribute. 
To match the 10-m resolution of the remote sensing inputs, we rasterize hedgerow lines through the following process, shown in Figure \ref{fig:hedgerows_downsampling}: we apply a buffer around each line to create 7m-wide hedgerows\footnote{7 meters is the average width of hedgerows measured across Germany \citep{drexler2024carbon}, and is the width used in hedgerow accounting calculations by the French government \citep{de2023haie}.}; rasterize buffered lines at a 0.5-m resolution (Figure \ref{fig:hedgerows_before}); and downsample the resulting binary raster image to the target 10-m resolution by averaging (Figure \ref{fig:hedgerows_after}, \ref{fig:hedgerows_satellite}). 

We split this national data into 1.28km x 1.28km square {\em patches}. 
This yields 603,564 data points (128x128 pixel images), which we eventually subsample to 2,995 for faster processing (\S\ref{sec:benchmark}).

To capture differences in vegetation type and agricultural systems, we assign an agriculturally-relevant climatic zone to each patch. We use data on Global Agro-Ecological Zones \citep{fao_gaez_v4}, available at a 10-km resolution, extract the thermal regime underlying the classification, and assign to each patch the regime that it overlaps the most with. We further group patches into a temperate zone (``Temperate, cool" and ``Temperate, moderately cool" regimes) and subtropic zone (``Subtropics, cool" and ``Subtropics, moderately cool" regimes), which comprise ${\sim}99\%$ of the data.

Finally, to evaluate whether the difficulty of the task varies inside and outside of agricultural zones, we use geospatial data on agricultural parcel boundaries across France \citep{cantelaube2014registre}.

%% file: text/benchmark.tex
The Hedgementation benchmark uses the hedgerow dataset to evaluate semantic segmentation approaches to hedgerow detection and assess model generalization across space and climatic zones.

\begin{figure}[t]
  \centering
  \makebox[\linewidth]{\hfill
  \begin{subfigure}{0.23\linewidth}
    \includegraphics[width=\linewidth]{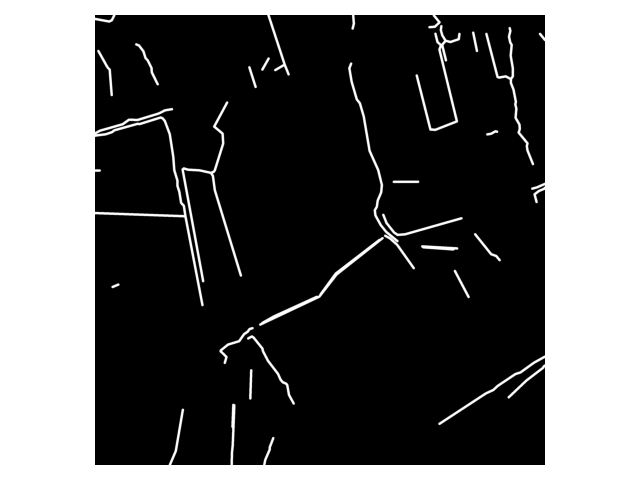}
    \caption{Hedgerows rasterized at 0.5-m resolution.}\label{fig:hedgerows_before}
  \end{subfigure}\hfill
  \begin{subfigure}{0.23\linewidth}
    \includegraphics[width=\linewidth]{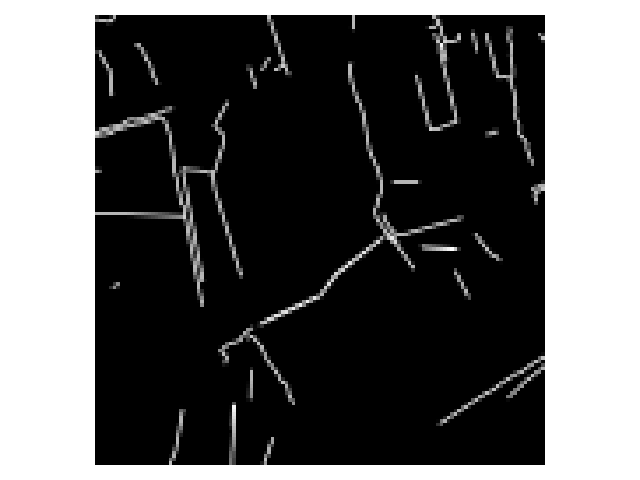}
    \caption{Final label raster at 10-m resolution.}\label{fig:hedgerows_after}
  \end{subfigure}\hfill
  \begin{subfigure}{0.23\linewidth}
    \includegraphics[width=\linewidth]{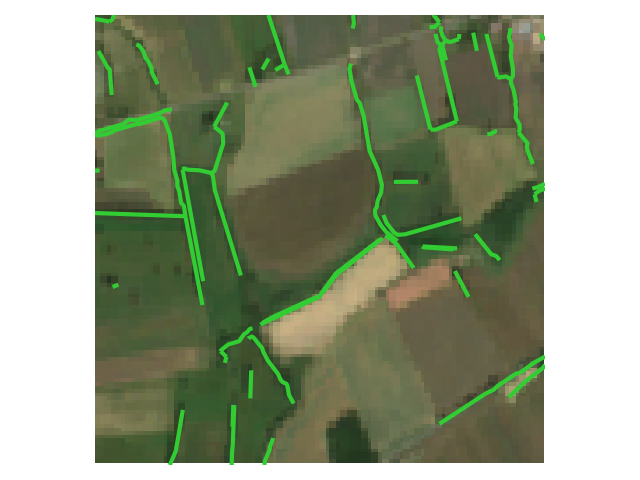}
    \caption{Final label raster overlaid on Sentinel-2 image.}\label{fig:hedgerows_satellite}
  \end{subfigure}\hfill
  \begin{subfigure}{0.23\linewidth}
    \includegraphics[width=\linewidth]{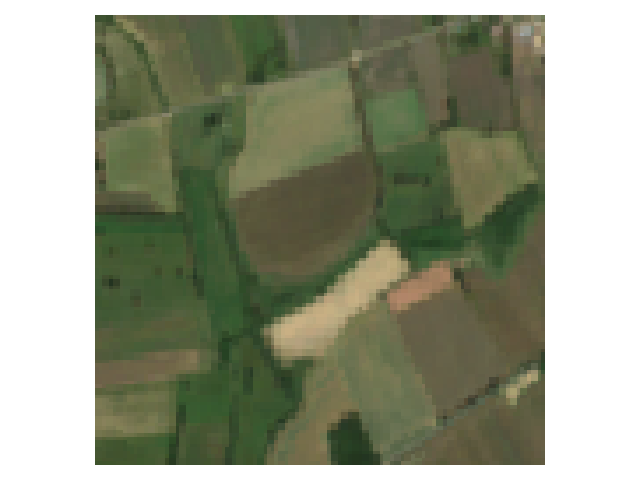}
    \caption{Sentinel-2 image for the chosen patch.}\label{fig:satellite_only}
  \end{subfigure}
  }%

  \caption{%
  Hedgerow label rasterization process, applied to one \emph{patch}: spatial detail is preserved.
  }
  \label{fig:hedgerows_downsampling}
  \vspace{-8pt}  
\end{figure}

\begin{figure}[t]
  \centering
  \makebox[\linewidth]{\hfill
  \begin{subfigure}{0.31\linewidth}
    \includegraphics[width=\linewidth]{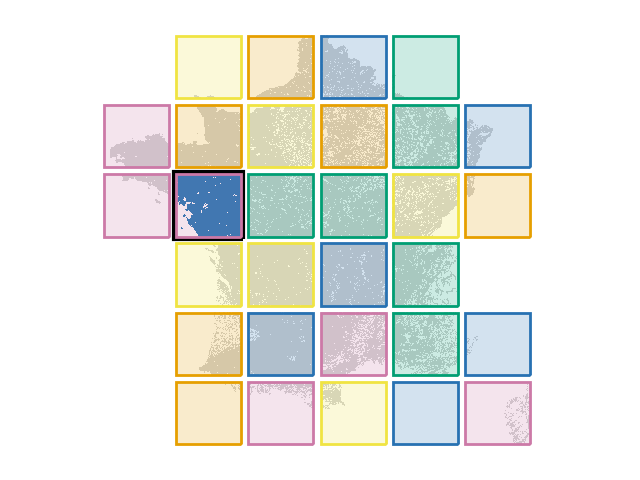}
    \caption{Tiles for testing spatial generalization. Empty tiles are dropped.}\label{fig:tiles}
  \end{subfigure}\hfill
  \begin{subfigure}{0.31\linewidth}
    \includegraphics[width=\linewidth]{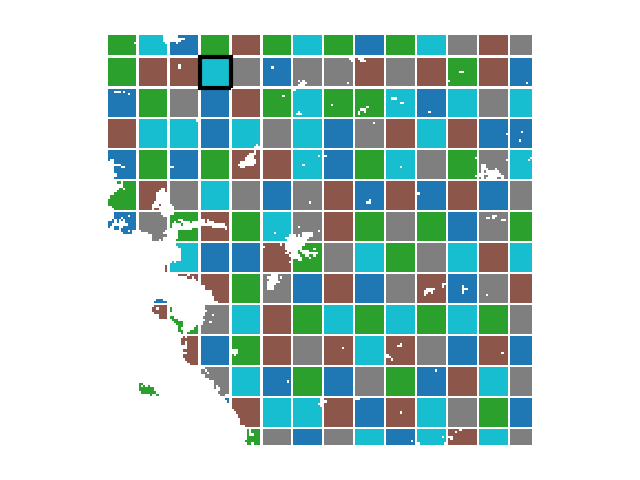}
    \caption{Cells with a spatial buffer, assigned to train/val/test sets.}\label{fig:cells}
  \end{subfigure}\hfill
  \begin{subfigure}{0.31\linewidth}
    \includegraphics[width=\linewidth]{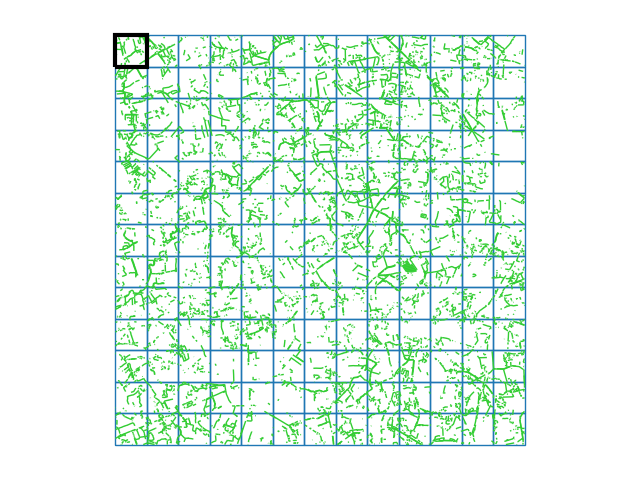}
    \caption{One cell of 1.28km x 1.28km patches (hedgerows are in green).}\label{fig:patches}
  \end{subfigure}
}%
  \caption{%
  Dataset creation process:
  (\ref{fig:tiles}) France is split into large tiles, randomly assigned to folds (shade colors) that determine near/far test sets for spatial generalization.
  (\ref{fig:cells}) Each tile is split into cells separated by a spatial buffer; cells are randomly assigned to train/val/test sets, via 5 folds (colors).
  (\ref{fig:patches}) Each cell is a 12x12 grid of 1.28km x 1.28km patches (individual data points).}
  \label{fig:dataset_creation}
  \vspace{-8pt}  
\end{figure}

\textbf{Assignment to train/val/test.}
We first group patches into cells of 12x12 patches 
separated by a 2-patch-wide buffer.
We then randomly assign cells into one of five folds (see Figure \ref{fig:cells}), and partition the folds into three for training (60\% of the data), one for validation (20\% of the data) and one for testing (20\% of the data).
The spatial buffer between cells ensures that no validation patch is directly adjacent to a training patch, and no test patch is directly adjacent to a training or validation patch. This is important as hedgerows and other landscape features can cross patch boundaries.

\textbf{Evaluation metric.}
To evaluate hedgerow detection performance, we binarize the continuous hedgerow labels ($y_i^r \in [0,1]$ for pixel $i$ from averaging higher-resolution binary pixels) into $y_i = 1$ iff $y_i^r > 0$.
We measure the quality of our models' binary predictions $\hat{y_i}$ by the Intersection over Union (IoU)~\citep{jaccard1901etude}, a standard metric for pixel-wise classification tasks
~\citep{shelhamer2016fully}.
$\text{IoU} = \frac{\sum_i (\hat{y}_i=1 \ \text{and} \ y_i=1)}{\sum_i(\hat{y}_i=1 \ \text{or} \ y_i=1)}$ with summation over all test pixels.
This metric is invariant to scale and sparsity, which is relevant to our setting, and is a common metric for remote sensing tasks~\citep{lacoste2024geo,kikaki2024detecting,garnot2021panoptic}.

\textbf{Spatial generalization.}
To assess spatial generalization, we split France into large {\em tiles} separated by a 28km buffer, as depicted in Figure \ref{fig:tiles}.
We drop empty tiles, and randomly assign the 30 remaining tiles to five folds.
Four folds provide train/val patches, and the test patches from those folds are potentially near training patches, thus constituting our {\em near test} set.
From the remaining fold, we only keep test patches, which constitute our {\em far test} set.
These test patches are at least 28km away from any training patch, and potentially much further.
We can compare model performance on test patches from the {\em near} vs. {\em far test} set.
We use cross-validation, with each fold serving as the {\em far} fold in turn, and document performance averages and variance across folds.

\textbf{Overall dataset size.}
We subsample a 600 patches per fold stratified over each tile, splitting folds 3/1/1 across train/val/test.
This creates our final dataset: 1,797 patches in train, 599 in valid, and 599 in test, totalling 2,995 patches. It contains 49,070,080 pixels, of which 4.95\% are hedgerow pixels. 

\textbf{Climate zone generalization.}
To measure performance variability and generalization across different types of landscapes, we evaluate models separately on test patches in the temperate zone and on test patches in the subtropical zone.
We also compare performance when training on temperate training patches only, subtropical training patches only, or both.
Climatic zones are unequally represented, with temperate patches outnumbering subtropical ones 3-to-1.
To remove confounding from training data size, we subsample all training sets for this analysis to the  subtropical training set size. 

%% file: text/results.tex
\textbf{Models.}
We explore two approaches for Hedgementation and fit five models.
First, we train deep learning models from scratch, using raw satellite imagery to predict hedgerows.
We fit models from two machine learning for remote sensing papers: the PASTIS U-net with Temporal Attention Encoder~\citep{garnot2021panoptic} (PASTIS U-TAE), and the Fields of The Worlds model~\citep{ftw} (FTW). 
Both models are trained from random initialization, though we experiment with fine-tuning from pre-training (Appendix \ref{appendix:space}).

We use the PASTIS U-TAE as-is, altering only the output layer for binary classification, with no changes to model architecture or hyperparameters.
The FTW model uses fewer spectral bands and only two input images from different times of the year. 
Appendix \ref{appendix:ftw-arch} details its alteration.

Second, we use the AEF embeddings
~\citep{aef} with simpler models.
These embeddings have been positively evaluated on vegetation and agriculture tasks~\citep{ma2025harvesting,houriez2025scalable}.
We use them with: a kNN enhanced with $3\times3$ spatial context via concatenated embeddings and center weighting to reduce noise, with a $200k$ balanced training set restricting positive samples to high-confidence interior hedgerow pixels; a Random Forest; and a Logistic Regression.
Details and hyper-parameter tuning are in Appendix \ref{appendix:space}.

\begin{table}[t]
\centering
\scriptsize
\setlength{\tabcolsep}{4pt}
\renewcommand{\arraystretch}{0.95}
\resizebox{\linewidth}{!}{%
\begin{tabular}{l cccccccccccc}
\toprule
& \multicolumn{2}{c}{G0} & \multicolumn{2}{c}{G1} & \multicolumn{2}{c}{G2} & \multicolumn{2}{c}{G3} & \multicolumn{2}{c}{G4} & \multicolumn{2}{c}{Avg} \\
\cmidrule(lr){2-3} \cmidrule(lr){4-5} \cmidrule(lr){6-7} \cmidrule(lr){8-9} \cmidrule(lr){10-11} \cmidrule(lr){12-13}
Method & Near & Far & Near & Far & Near & Far & Near & Far & Near & Far & Near & Far \\
\midrule
PASTIS U-TAE        & 0.409 & 0.417 & 0.408 & 0.392 & 0.416 & 0.341 & 0.425 & 0.374 & 0.393 & 0.420 & \textbf{0.4102} & \textbf{0.3888} \\
FTW Model           & 0.32  & 0.35  & 0.31  & 0.34  & 0.30  & 0.22  & 0.30  & 0.29  & 0.31  & 0.27  & \textbf{0.31}   & \textbf{0.29}   \\
AEF: Random Forest  & 0.236 & 0.228 & 0.232 & 0.245 & 0.243 & 0.212 & 0.231 & 0.249 & 0.232 & 0.249 & \textbf{0.235}  & \textbf{0.237}  \\
AEF: Log. Reg.      & 0.206 & 0.198 & 0.202 & 0.214 & 0.211 & 0.172 & 0.200 & 0.217 & 0.209 & 0.193 & \textbf{0.206}  & \textbf{0.200}  \\
AEF: KNN ($k$=3)    & 0.155 & 0.134 & 0.147 & 0.158 & 0.155 & 0.119 & 0.145 & 0.173 & 0.155 & 0.136 & \textbf{0.151}  & \textbf{0.157}  \\
\bottomrule
\end{tabular}%
}
\caption{IoU (higher = more accurate) on our Hedgementation benchmark across groups (G0, $\dots$).}
\label{tab:iou_all}
\end{table}

\textbf{Analysis: Generalization over space.}
Table \ref{tab:iou_all} compares baseline performance on the near and far test sets.
Figure \ref{fig:qual_grid_inputs_preds} and Appendix \ref{sec:results-visual} show qualitative predictions on four patches.
On average, the PASTIS and FTW models are better on nearer patches than on further ones (+0.02 IoU/more than 5\% higher), as expected.
kNN performs similarly on near and far. 
This might be due to the global training of the embedding. 
However, performance is weak overall (see Figure \ref{fig:qual_grid_inputs_preds}).
The difference due to distance is minor, with significant variation across near/far tile groupings, with (G0, G4) yielding better performance on the far set.
We thus turn to generalization over climatic zones.

\begin{table}[t]
\centering
\small
\setlength{\tabcolsep}{3pt}
\renewcommand{\arraystretch}{1.15}
\newcommand{\twotablescale}{0.85}
\noindent
\begin{minipage}[t]{0.32\linewidth}
\vspace{0pt}\centering
\scalebox{\twotablescale}{%
\begin{tabular}{l l c}
\hline
\textbf{Trained on} & \textbf{Tested on} & \textbf{IoU} \\
\hline
Full dataset      & Full dataset      & 0.416 \\
Full dataset      & Temp. only        & 0.424 \\
Full dataset      & Subtropic only    & 0.374 \\
\hline
Subs. dataset     & Full dataset      & 0.368 \\
Subs. dataset     & Temp. only        & 0.380 \\
Subs. dataset     & Subtropic only    & 0.316 \\
\hline
\end{tabular}%
}
\end{minipage}\hfill
\begin{minipage}[t]{0.32\linewidth}
\vspace{0pt}\centering
\scalebox{\twotablescale}{%
\begin{tabular}{l l c}
\hline
\textbf{Trained on} & \textbf{Tested on} & \textbf{IoU} \\
\hline
Subtropic only    & Full dataset      & 0.352 \\
Subtropic only    & Temp. only        & 0.357 \\
Subtropic only    & Subtropic only    & 0.331 \\
\hline
Temperate only    & Full dataset      & 0.357 \\
Temperate only    & Temp. only        & 0.372 \\
Temperate only    & Subtropic only    & 0.287 \\
\hline
\end{tabular}%
}
\end{minipage}\hfill
\begin{minipage}[t]{0.32\linewidth}
\vspace{0pt}\centering
\scalebox{\twotablescale}{%
\begin{tabular}{l l c}
\hline
\textbf{Trained on} & \textbf{Tested on} & \textbf{IoU} \\
\hline
All pix (subs.) & All pix        & 0.410 \\
All pix (subs.) & Agri. pix      & 0.459 \\
All pix (subs.) & Non-Agri. pix  & 0.297 \\
\hline
Agri. pix       & All pix        & 0.378 \\
Agri. pix       & Agri. pix      & 0.457 \\
Agri. pix       & Non-Agri. pix  & 0.243 \\
\hline
\end{tabular}%
}
\end{minipage}
\caption{PASTIS U-TAE generalization on climatic zones (left, center) and agricultural zones (right).
}
\label{tab:pastis_iou_splits}
\vspace{-8pt}
\end{table}

\textbf{Analysis: Generalization over climatic zones.}
Table \ref{tab:pastis_iou_splits} focuses on our best model (PASTIS U-TAE) to compare generalization across climate zones.
Subtropical patches are consistently more difficult to predict, even for the model trained exclusively on these patches.
The smallest gap is for the subtropic-only model (2.1\%), and the largest gap is for the temperate-only model (7.0\%). 
Interestingly, while training on a random subset does worse on subtropic tiles than the subtropic-only model, it does better on temperate tiles than the temperate-only model.
We confirm that dataset size matters: the model trained on less data is much worse than the model trained on all the data.

\textbf{Analysis: Performance in agricultural zones.}
Hedgerows are prevalent and particularly important in agricultural regions, where BD Haies label quality is also higher thanks to validation by individual farmers.
We find that training on all pixels yields better results than restricting training to agricultural zones (Table \ref{tab:pastis_iou_splits}). However, predicting hedgerows in agricultural landscapes is significantly easier: the IoU of the PASTIS U-TAE model reaches 45.9\% of IoU on agricultural pixels, up from 41\% on all pixels.
This is a valuable and encouraging finding for applications that focus on these landscapes.

%% file: text/discussion-short.tex
\textbf{Discussion: usage and limitations.}
The hedgerow task is complementary to agricultural tasks such as field boundary mapping~\citep{ftw} and crop type mapping~\citep{pastis,russwurm2019breizhcrops,worldcerealsbenchmark}.
It is a spatial challenge due to the thin extents, diverse aspects, and various arrangements of hedgerows, and a generalization challenge due to the variations in the surrounding landscapes (complexity, tree density) across climate / agro-ecological zones. 

The Hedgementation imagery and labels are derived from free and public data sources, for accessibility and reproducibility, and extensive quality checks support its labeling.
As a benchmark with ground truth on the ground, it complements benchmarks with image-derived labels~\citep{helber2019eurosat}.
We discuss additional related work, limitations, and next steps in~\autoref{appendix:discussion}.
While there are improvements to be made, this version of Hedgementation can already inform model development with its applied task and measures of spatial and agricultural generalization.

%% file: text/appendix/app-data.tex
Labels come from a nationwide inventory of hedgerows in France that was recently updated.
The original inventory, ``BD Haies V1'', is obtained by aggregation and linearization of two data sources: (1) hedgerows identified in agricultural landscapes as part of compulsory reporting for the \emph{Registre Parcellaire Graphique} (RPG), which covers more than 90\% of the agriculture zone \citep{cantelaube2014registre}--- identification of these hedgerows uses pre-processed aerial photographs, dating from 2011-2014 depending on the region, and validated by individual farmers; and (2) hedgerows identified by automatic image segmentation and photo interpretation, as part of the vegetation layer of the national topographic database BD TOPO---identification of these hedgerows uses imagery from 2004-2015 depending on the region, is not restricted to the agricultural zone, but is restricted to longer (25m+) and higher (1,30m+) hedgerows \citep{IGN_OFB_2020_BDHaie}.
BD Haies V1 is thus representative of the status of hedgerows around 2004-2014.
The updated version, ``BD Haies V2'', used in this paper, is constructed by revision of BD Haies V1 and involves two main data sources: (1) hedgerows identified in agricultural landscapes as part of more recent iterations of the RPG, now involving imagery from 2020-2022 depending on the region; and (2) numerical surface models from corresponding years that serve to further validate the continued existence of V1 hedgerows \citep{Commagnac_2024_BDHaieV2}.
BD Haies V2 is thus representative of the status of hedgerows around 2020-2022.

%% file: text/appendix/app-ftw.tex
FTW uses input satellite imagery in a slightly different format than PASTIS, such that it cannot ingest as is the data provided by Hedgementation. We pre-process input data as follows for use by the FTW model:

\begin{itemize}
    \item \textbf{Scene date selection:} FTW uses imagery from two temporal windows per geographical zone, corresponding to the planting and harvest seasons, with dates adapted to each country’s agricultural calendar (e.g., in France, March–June for window A and August–September for window B). Basde on the Sentinel-2 imagery extracted for Hedgementation, each window contains between 30 and 60 satellite images. For each window (A and B), a single representative image is selected based on the median acquisition date within the interval, reducing sensitivity to irregular acquisitions and transient noise. 
    \item \textbf{Cloud coverage:} No modification is applied.
    \item \textbf{Spatial dimensions:} The original spatial dimensions of the generated dataset are preserved.
    \item \textbf{Spatial resolution:} No modification is applied.
    \item \textbf{Spectral bands:} Only the B04, B03, B02, and B08 bands are extracted from the generated dataset patches and stacked following the same procedure used in FTW.
\end{itemize}

\begin{figure}[H]
    \centering
    \includegraphics[width=1\linewidth]{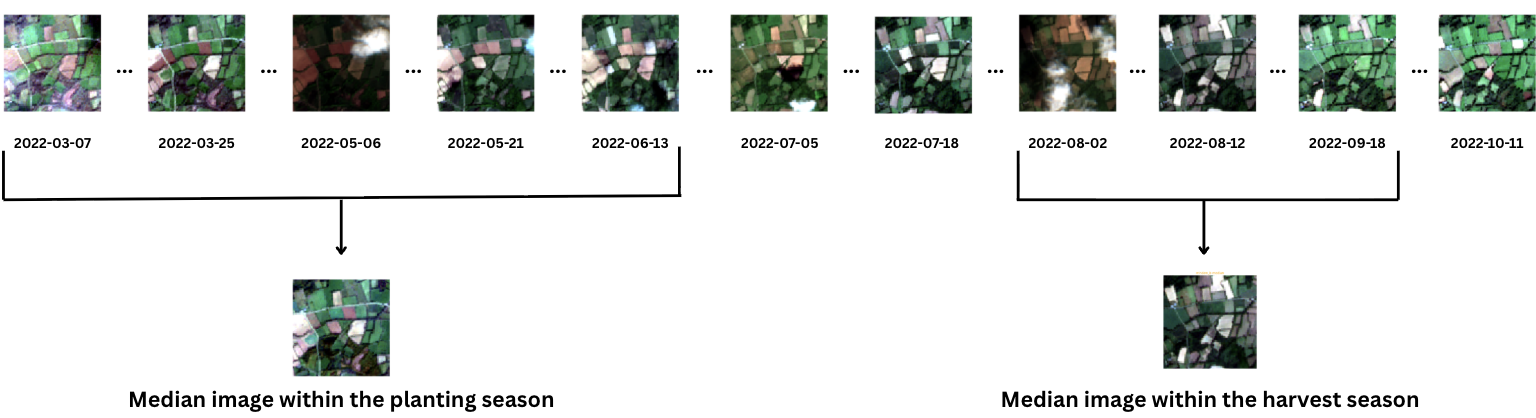}
    \caption{%
    Input data for the FTW model: the model requires two input images taken from two characteristic times of year.
    We collect the necessary imagery for the model from the image time-series collected for our Hedgementation dataset.
    }
    \label{fig:ftw_input}
\end{figure}

Figure~\ref{fig:ftw_input} illustrates the FTW input data processing. Window A corresponds to the median composite of Sentinel-2 images acquired during the planting season (March–June 2022), while Window B corresponds to the median composite of images acquired during the harvest season (August–October 2022). Each window is represented by a four-band multispectral image covering the same agricultural area.

%% file: text/discussion-long.tex
\textbf{Related Work.}
Pixel-wise prediction tasks, like our hedgerow task, are valuable for benchmarking not only recognition (what) but localization (where). 
As evidence of their importance, the popular GEO-Bench suite of tasks includes both image classification and pixel classification~\citep{lacoste2024geo}.
Pixel-wise benchmarks are likewise included in the comparisons of pre-trained models for remote sensing (or ``foundation models'')~\citep{cong2022satmae,tseng2025galileo,astruc2025anysat,aef,waldmann2025panopticon,jakubik2025terramind}.

Systematic hedgerow detection is not a new goal, and capacities have increased with the availability of remote sensing and machine learning tools. Yet, relatively few papers propose and evaluate approaches to map hedgerows at regional scales where data does not yet exist, in part due to limitations in label data or resource availability. One such example, \cite{ha2019shelterbelt}, use Sentinel imagery and object-based classification to map shelterbelts across Saskatchewan in 2016. Another, \cite{muro2025hedgerow}, use high-resolution PlanetScope imagery and semantic segmentation (standard U-net with a ResNet backbone) to map hedgerows across one state in Germany and evaluate transferability to other states. Related initiatives exist to map wooded landscape features more broadly. In particular, the Copernicus Land Monitoring Service produces the Small Woody Features data layer that spans much of Europe and includes hedgerows, though accuracy of hedgerow detection specifically is not known \citep{faucqueur2019new}. And \cite{liu2023overlooked} map all trees outside forests in Europe and evaluates transferability to boreal and temperate zones of North America, using PlanetScope imagery and semantic segmentation (U-Net with and EfficientNet-B4 backbone), though LiDAR-derived height data are used in lieu of annotations. A key limitation is that performance is not comparable across models, and workflows are generally not reproducible, which limits expansion of modeling approaches beyond their original spatial and temporal extent.

Our task is complementary to other agricultural segmentation tasks that focus on groups of points, like the recognition of crop types and fields~\cite{garnot2021panoptic}, because hedgerow segmentation focuses on lines.
In this way it is more related to the task of mapping field boundaries~\citep{ftw}.
At the same time, the hedgerow task is distinct: not all hedgerows are field boundaries, and not all field boundaries are hedgerows.

\textbf{Limitations and Next Steps.}

\emph{Label Quality and Consistency.}
The BD Haie data from which we derive our benchmark labels is intended to be exhaustive and accurate.
However, it may still be subject to noise and inconsistency, especially due to differences in hedgerow definition and in the detection process inside and outside of agricultural areas, use of different years of imagery across the country, changes in methodology across years, and lack of independent accuracy assessment.
To verify the quality and consistency of annotations, we will examine higher-resolution imagery provided by Planet~\cite{planet2022planet} and check for label noise by monitoring for outliers in the training loss.

\emph{Scope in Space, Time, and Modality.}
Hedgementation covers the one country of France, in the one time period of 2021–2022, and imagery from one satellite mission and optical modality: Sentinel-2.
We will next extend the benchmark in geography in modality.
For geography, we will propagate then verify labels in related agricultural regions of the UK and Canada~\citep{houriez2025scalable}.
For modality, we will combine our optical imagery with an additional modality such as SAR or LIDAR, as they are sensitive to height.
These sensors have been incorporated into models for recognizing tall and short crops~\citep{ditommaso2021combining} and could help differentiate hedgerows from other landscape features like fences.
Furthermore multi-modal data has been shown to help with data efficiency and generalization~\citep{rao2025using}, and we have already confirmed that generalization across agro-ecological zones is a challenge.

\emph{Prediction Quality.}
The accuracies achieved by our baselines are promising but insufficient for agricultural and ecological analysis.
We will strive to improve performance by transfer learning from supervised pre-training~\citep{pastis,bastani2023satlaspretrain} or self-supervised pre-training~\citep{herzog2025olmoearth}, more sophisticated sampling for curriculum learning~\citep{mindermann2022prioritized} and countering imbalance, and further tuning of the architecture and optimization.

%% file: text/visual-results.tex
\begin{figure}[t]
  \centering
  \includegraphics[width=0.24\linewidth]{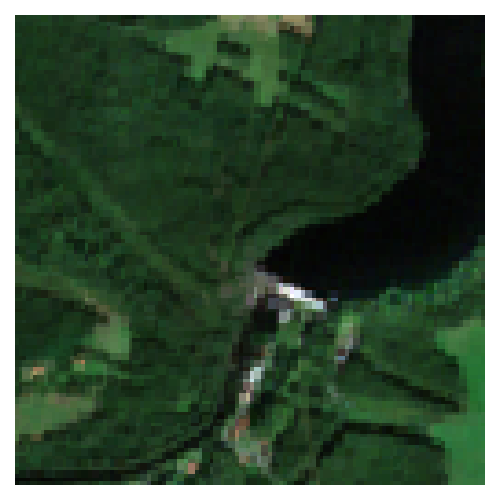}\hfill
  \includegraphics[width=0.24\linewidth]{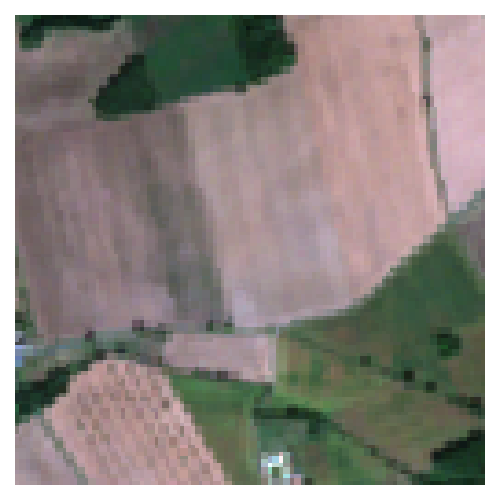}\hfill
  \includegraphics[width=0.24\linewidth]{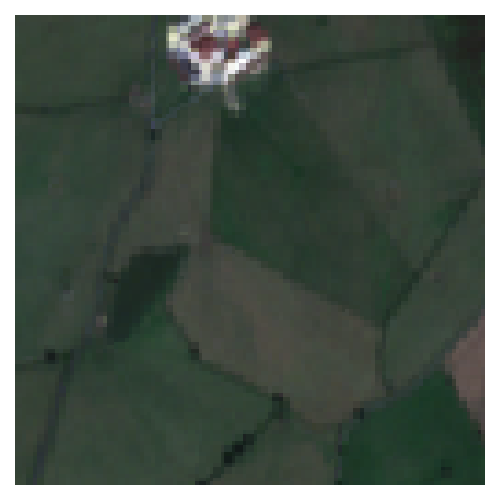}\hfill
  \includegraphics[width=0.24\linewidth]{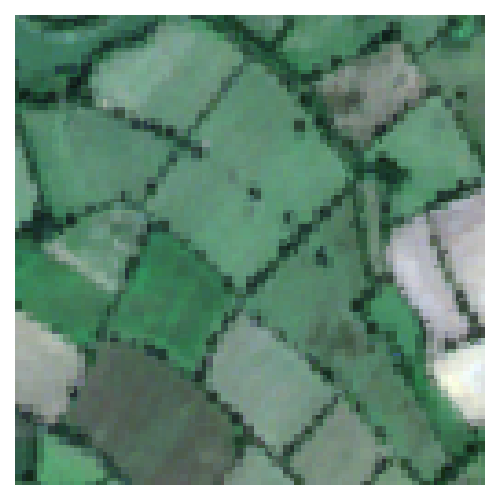}

  \vspace{0.1em}
  \includegraphics[width=0.24\linewidth]{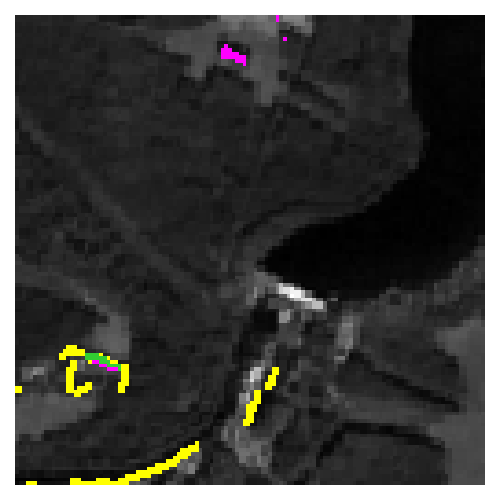}\hfill
  \includegraphics[width=0.24\linewidth]{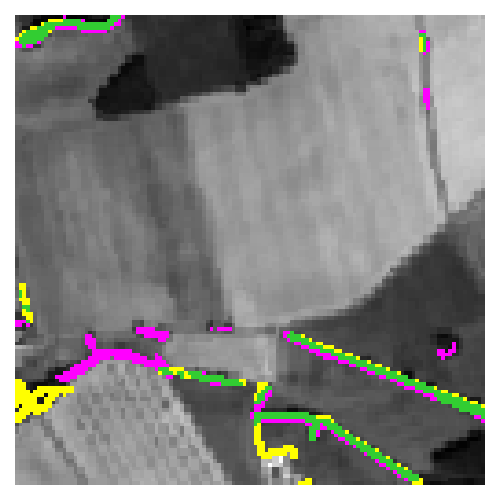}\hfill
  \includegraphics[width=0.24\linewidth]{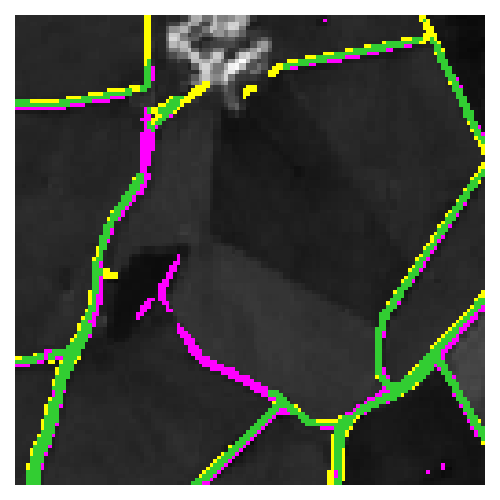}\hfill
  \includegraphics[width=0.24\linewidth]{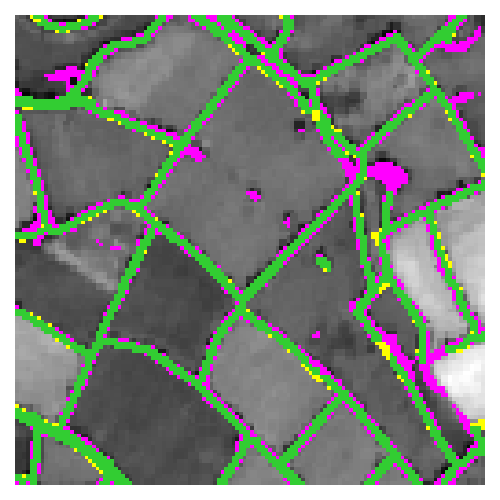}

    \vspace{0.1em}
  \includegraphics[width=0.24\linewidth]{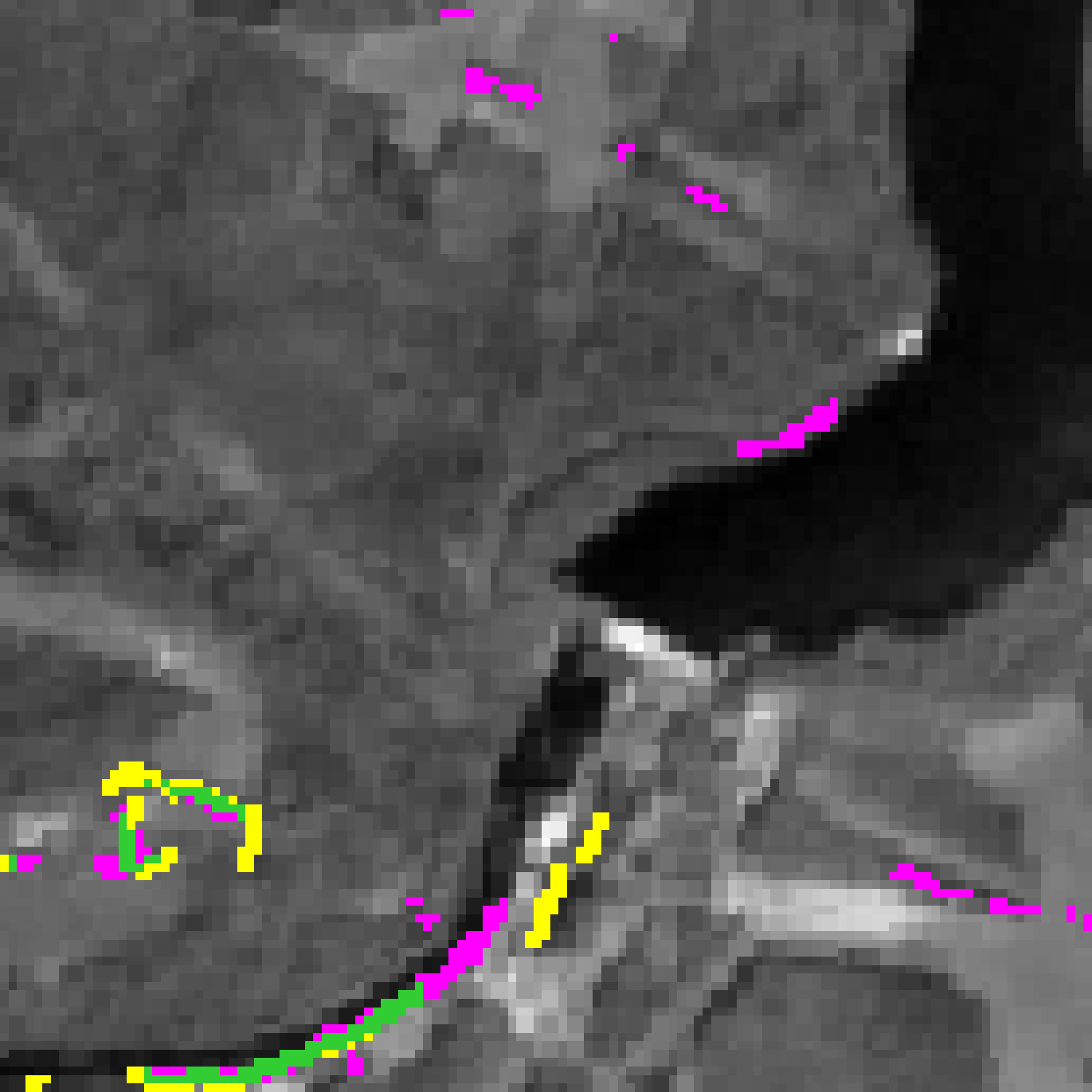}\hfill
  \includegraphics[width=0.24\linewidth]{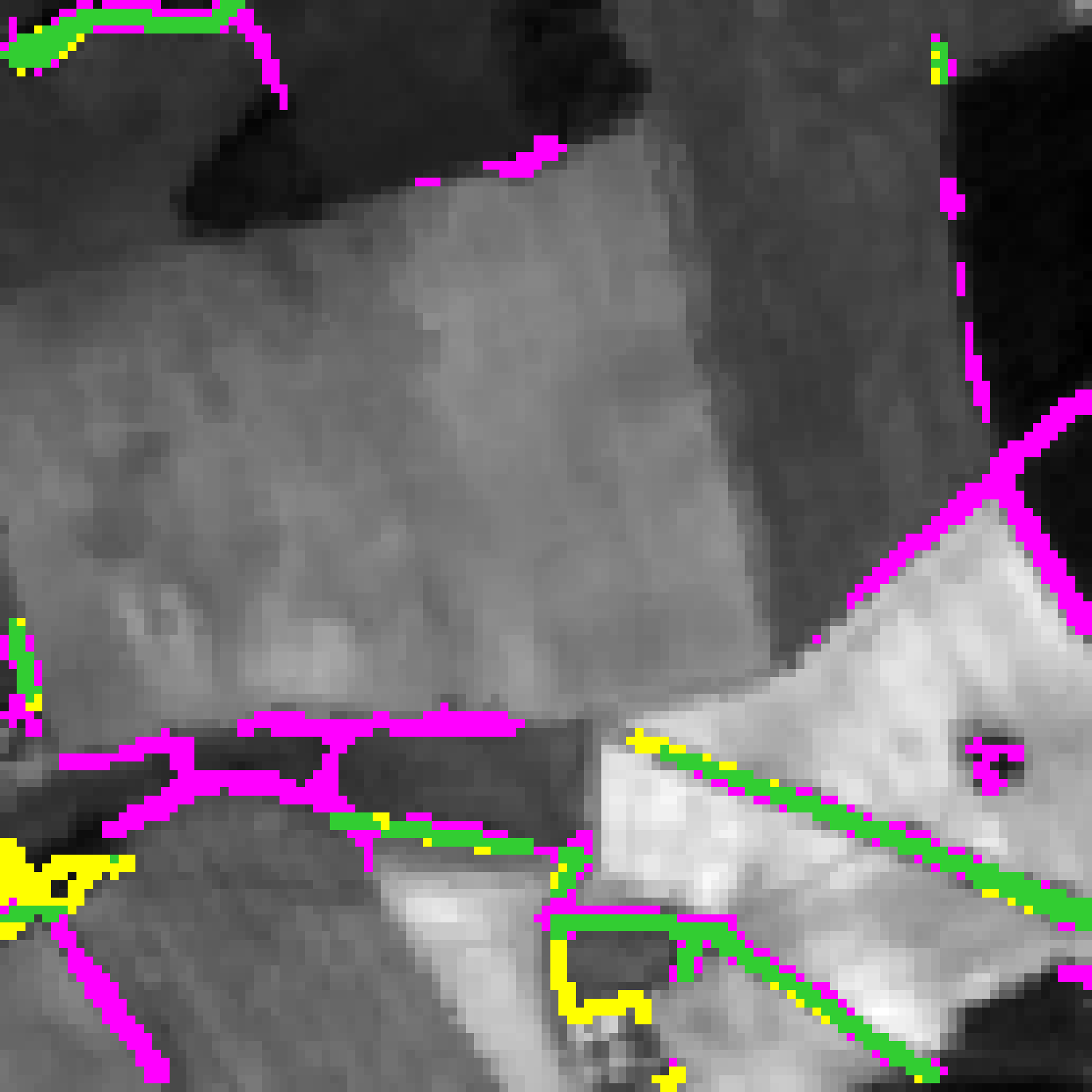}\hfill
  \includegraphics[width=0.24\linewidth]{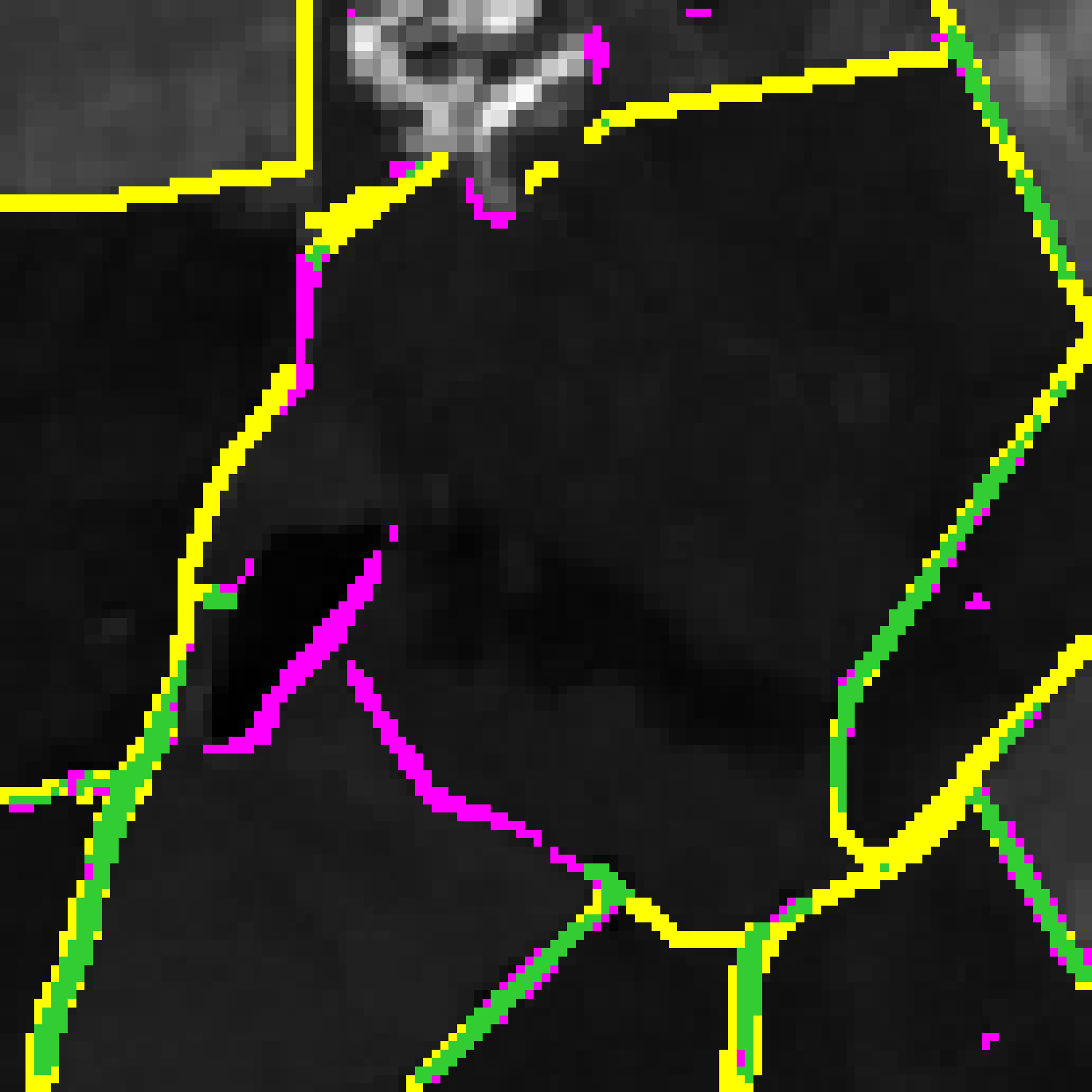}\hfill
  \includegraphics[width=0.24\linewidth]{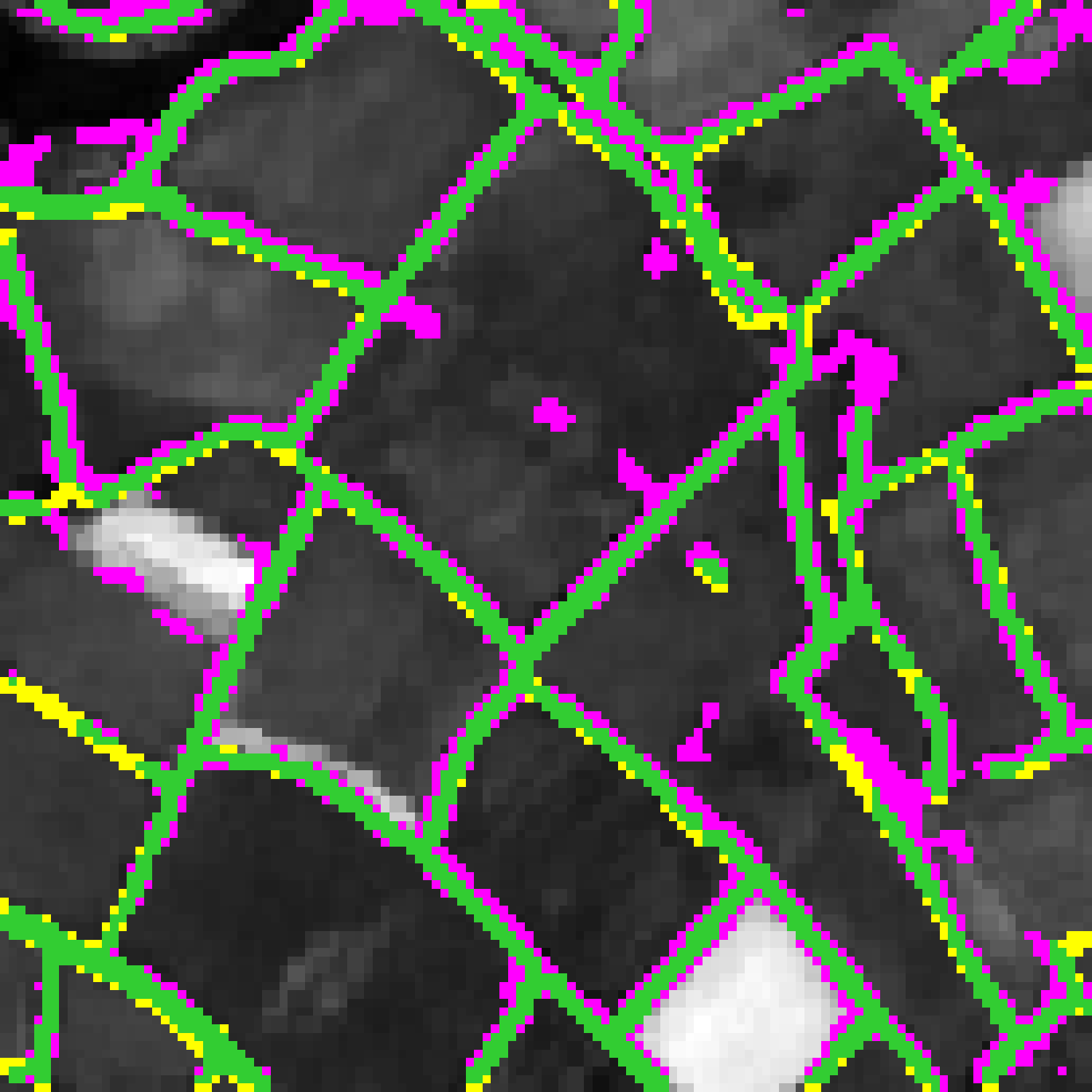}
  
  \vspace{0.1em}
  \includegraphics[width=0.24\linewidth]{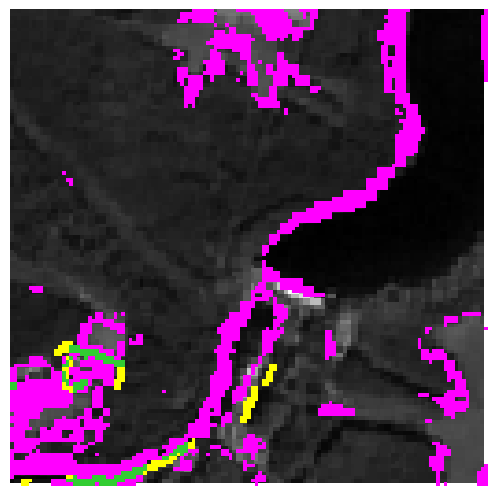}\hfill
  \includegraphics[width=0.24\linewidth]{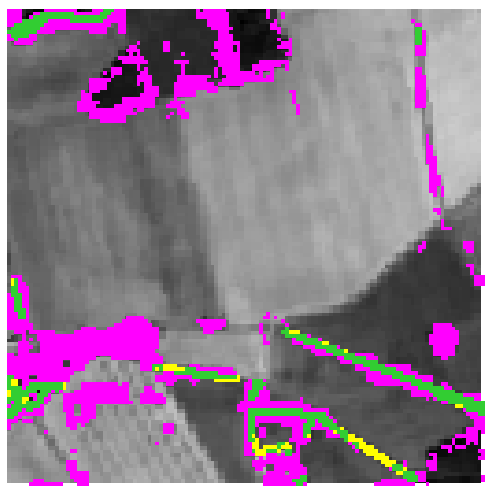}\hfill
  \includegraphics[width=0.24\linewidth]{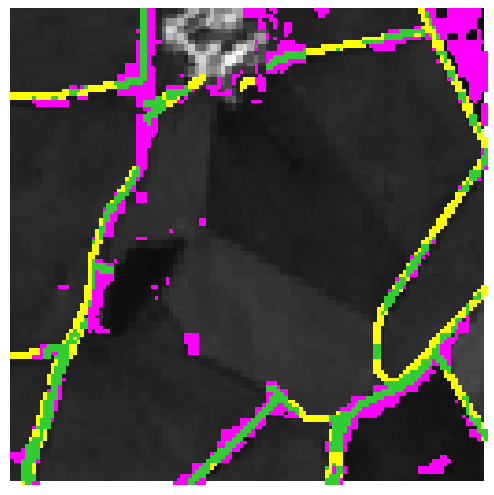}\hfill
  \includegraphics[width=0.24\linewidth]{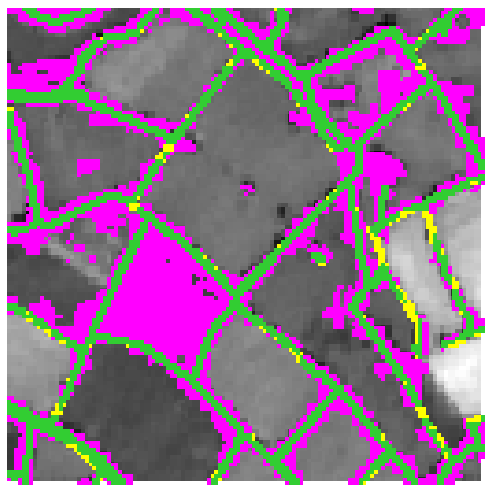}\hfill

  \vspace{0.1em}
  \includegraphics[width=0.24\linewidth]{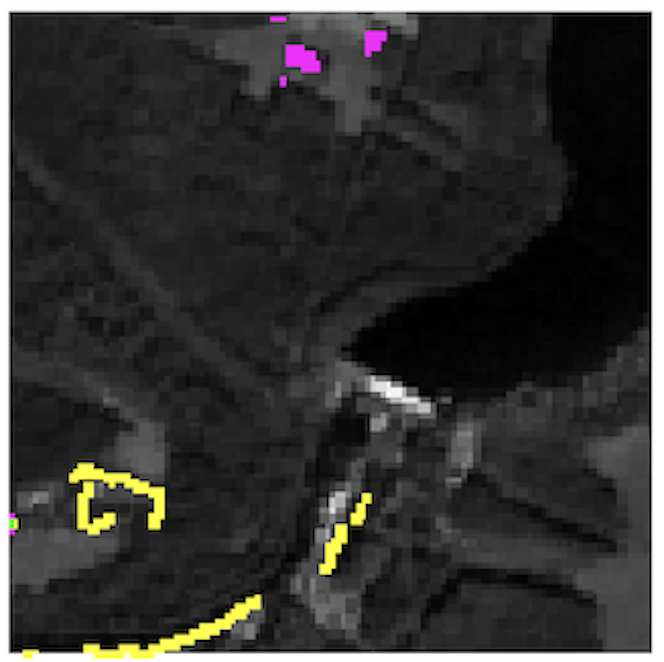}\hfill
  \includegraphics[width=0.24\linewidth]{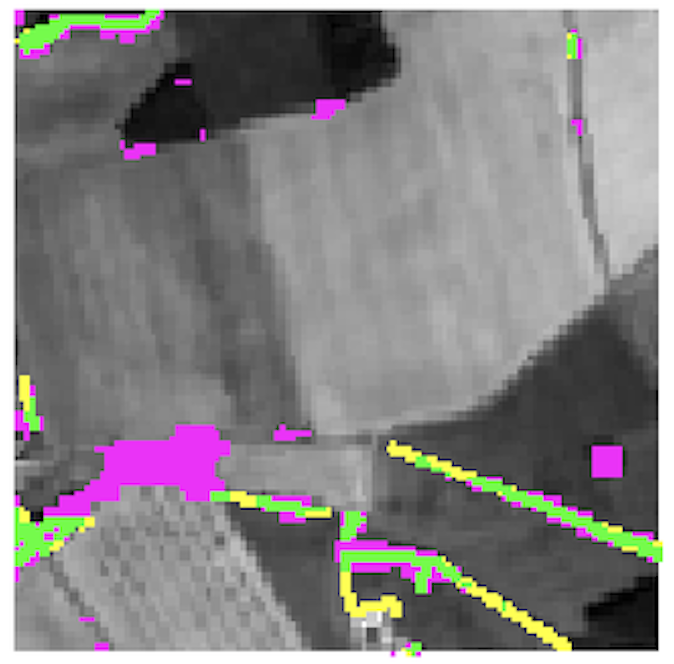}\hfill
  \includegraphics[width=0.24\linewidth]{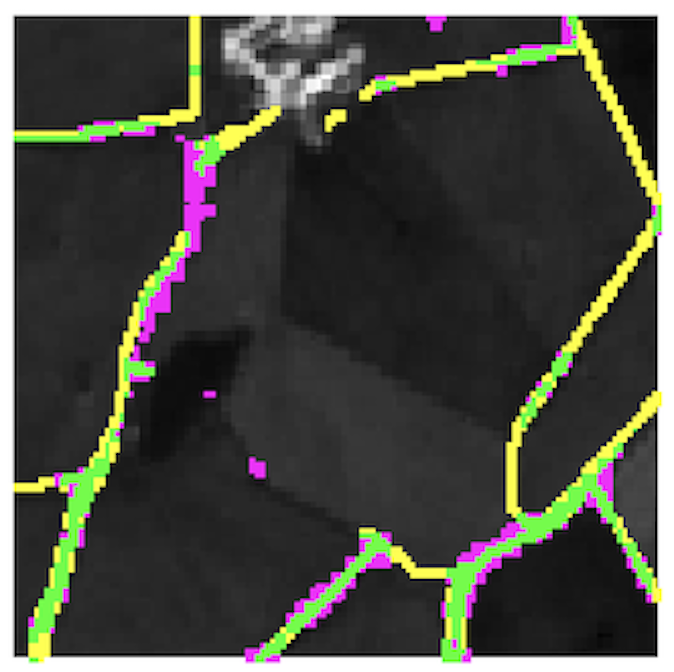}\hfill
  \includegraphics[width=0.24\linewidth]{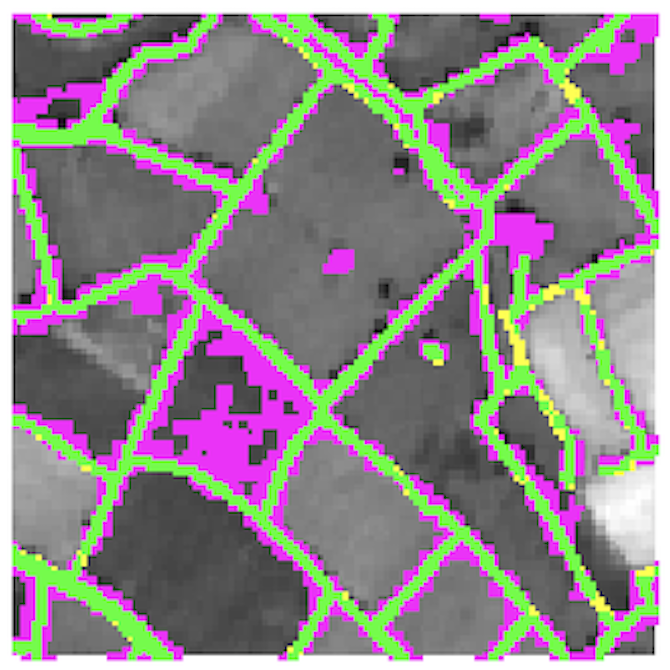}\hfill

\vspace{0.1em}
  \includegraphics[width=0.24\linewidth]{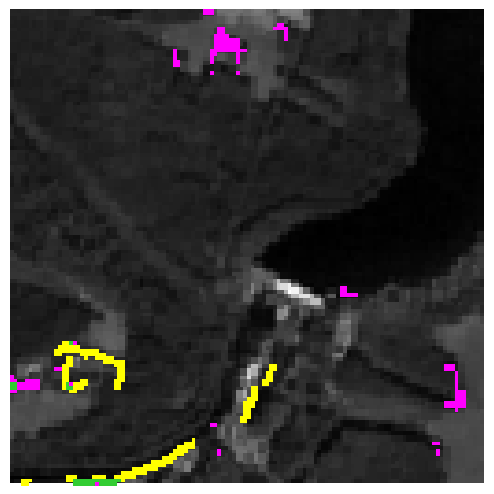}\hfill
  \includegraphics[width=0.24\linewidth]{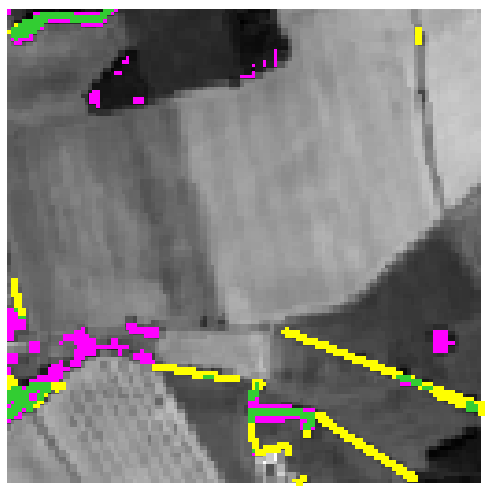}\hfill
  \includegraphics[width=0.24\linewidth]{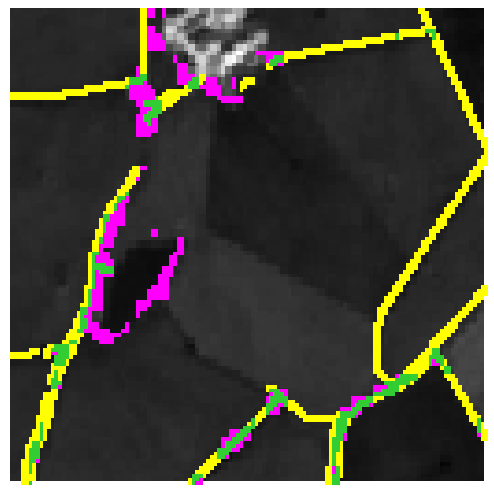}\hfill
  \includegraphics[width=0.24\linewidth]{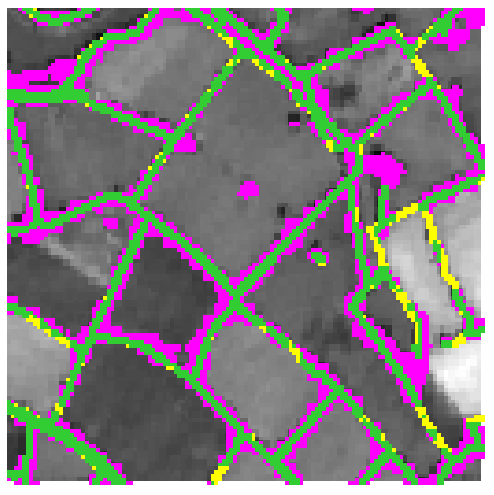}\hfill
  \caption{Qualitative examples of predictions for our best model in each class. We show input RGB patches (top), and predictions with TP=green, FN=yellow, and FP=magenta on following rows, in order for PASTIS, FTW, kNN (k=3), Random Forest and Logistic Regression.}
  \label{fig:qual_grid_inputs_preds}
\end{figure}

Figure \ref{fig:qual_grid_inputs_preds} shows four patches (first row) and predictions from the PASTIS, FTW, kNN, Random Forest, and Logistic Regression models on those in rows 2, 3, 4, 5, and 6 respectively. True Positives are shown in green, False Negatives are shown in yellow, and False Positives are shown in magenta.

%% file: text/appendix/results_pastis_space.tex
\begin{table}[H]
\centering
\begin{tabular}{|c|c|c|c|c|c|c|c|c|}
\hline
\multicolumn{9}{|c|}{\textbf{PASTIS UTAE on Hedgementation}} \\
\hline
\textbf{Experiment} & \multicolumn{4}{c|}{\textbf{Test(Near)}} & \multicolumn{4}{c|}{\textbf{Test(Far)}} \\
\hline
\textbf{Far Group} & \textbf{IoU} & \textbf{Precision} & \textbf{Recall} & \textbf{F1} & \textbf{IoU} & \textbf{Precision} & \textbf{Recall} & \textbf{F1} \\
\hline
G0 & 0.409 & 0.529 & 0.643 & 0.58 & 0.417 & 0.524 & 0.67 & 0.588 \\
\hline
G1 & 0.408 & 0.526 & 0.645 & 0.579 & 0.392 & 0.48 & 0.683 & 0.563 \\
\hline
G2 & 0.416 & 0.544 & 0.638 & 0.587 & 0.341 & 0.551 & 0.473 & 0.509 \\
\hline
G3 & 0.425 & 0.537 & 0.672 & 0.597 & 0.374 & 0.499 & 0.599 & 0.544 \\
\hline
G4 & 0.393 & 0.523 & 0.613 & 0.565 & 0.42 & 0.568 & 0.617 & 0.591 \\
\hline
\textbf{Avg} & \textbf{0.4102} & \textbf{0.5318} & \textbf{0.6422} & \textbf{0.5816} & \textbf{0.3888} & \textbf{0.5244} & \textbf{0.6084} & \textbf{0.559} \\
\hline
\end{tabular}
\caption{Results on the Far and Near test sets, when trained on the Near train set, for all tile-folds.}
\label{tab:pastis_space}
\end{table}

Table \ref{tab:pastis_space} shows the performance of the PASTIS U-TAE model on our Hedgementation benchmark. Focusing on the IoU, our prefered metric, we see that mean IoU is lower when evaluating on the far group than the near group with a 0.021\% gap. However, this trend is less consistent at the level of individual folds and models. Two models, G0 and G4, actually perform better on their far group than on the near group. Models G1, G2 and G3 all perform worse on their near group, by 1.6\%, 7.5\%, and 5.1\% IoU respectively.

\begin{table}[H]
\centering
\begin{tabular}{|l|l|c|c|c|c|c|}
\hline
\textbf{Model Architecture} & \textbf{Weights} & \textbf{Layers Trained} & \textbf{IoU} & \textbf{Precision} & \textbf{Recall} & \textbf{F1} \\
\hline
PASTIS U-TAE & Random Init & All & 0.415 & 0.533 & 0.651 & 0.586 \\
\hline

PASTIS U-TAE & PASTIS & All & 0.414 & 0.551 & 0.624 & 0.585 \\
\hline

PASTIS U-TAE & PASTIS & out\_conv  & 0.081 & 0.242 & 0.108 & 0.149 \\

\hline
\end{tabular}
\caption{Pretrained vs. Randomly Initialized Weights}
\label{tab:finetune}
\end{table}

In addition, we experiment with fine-tuning the weights of the original PASTIS model for our semantic segmentation task. We test two approaches: fine-tuning all the layers of the model; and fine-tuning the final convolutional block. All hyperparameters except for the initial weights were shared with our other experiments.
Neither approach yield an improvement over training from randomly initialized weights. Training all layers performs almost exactly on-par with the results of training from scratch, while training only the final convolutional block yields significantly inferior results. Additional experiments to explore this further are an interesting avenue for future work.

%% file: text/appendix/results_ftw_space.tex
Table \ref{table:ftw_near_far} shows the results of run the FTW model on our extrapolation over space task.
We observe that the far group is slightly harder to predict on average, but the difference is not very large. The performance of this model is lower than that of our best model, the PASTIS U-TAE \S\ref{appendix:pastis-results-space}.

\begin{table}[H]
\centering
\begin{tabular}{|c|c|c|c|c|c|c|c|c|}
\hline
\multicolumn{9}{|c|}{\textbf{FTW on Hedgementation}} \\
\hline
\textbf{Experiment} & \multicolumn{4}{c|}{\textbf{Test (Near)}} & \multicolumn{4}{c|}{\textbf{Test (Far)}} \\
\hline
\textbf{Far Group} & \textbf{IoU} & \textbf{Precision} & \textbf{Recall} & \textbf{F1} & \textbf{IoU} & \textbf{Precision} & \textbf{Recall} & \textbf{F1} \\
\hline
G0 & 0.32 & 0.67 & 0.38 & 0.49 & 0.35 & 0.65 & 0.42 & 0.51 \\
\hline
G1 & 0.31 & 0.60 & 0.39 & 0.47 & 0.34 & 0.59 & 0.44 & 0.50 \\
\hline
G2 & 0.30 & 0.70 & 0.34 & 0.46 & 0.22 & 0.67 & 0.25 & 0.36 \\
\hline
G3 & 0.30 & 0.69 & 0.35 & 0.46 & 0.29 & 0.74 & 0.33 & 0.45 \\
\hline
G4 & 0.31 & 0.66 & 0.38 & 0.48 & 0.27 & 0.65 & 0.31 & 0.42 \\
\hline
\textbf{Avg} & \textbf{0.31} & \textbf{0.66} & \textbf{0.37} & \textbf{0.47} & \textbf{0.29} & \textbf{0.66} & \textbf{0.35} & \textbf{0.45} \\
\hline
\end{tabular}

\caption{Results by far group on Hedgementation (Near vs. Far test sets).}\label{table:ftw_near_far}
\end{table}

Fine-tuning however can help the FTW model: as we see in Table~\ref{table:finetune-ftw}, initializing with ImageNet weights provides a modest but consistent improvement over random initialization, with IoU increasing from 0.333 to 0.359 and F1 from 0.500 to 0.529. This suggests that low-level visual features learned on natural images transfer usefully to satellite imagery even across this domain gap.

\begin{table}[H]
\centering
\begin{tabular}{|l|l|c|c|c|c|c|}
\hline
\textbf{Model Architecture} & \textbf{Weights} & \textbf{Layers Trained} & \textbf{IoU} & \textbf{Precision} & \textbf{Recall} & \textbf{F1} \\
\hline
FTW & Random Init & All & 0.333 & 0.607 & 0.425 & 0.500 \\
\hline
FTW & ImageNet & All & 0.359 & 0.575 & 0.489 & 0.529 \\
\hline
\end{tabular}
\caption{Pretrained vs. Randomly Initialized Weights for FTW}
\label{table:finetune-ftw}
\end{table}

%% file: text/appendix/results_knn_space.tex
We use pre-trained, publicly released AEF embeddings~\cite{aef} to train a k-Nearest Neighbours (kNN) model for the hedgerow detection task. We tune the number of neighbours $k \in \{1,3,5,\dots,25\}$ and the training set size, constructed by randomly sampling a balanced set of hedgerow and non-hedgerow pixels from the training data. We explore training set sizes ranging from very small samples (e.g., 1 positive / 1 negative) to larger balanced subsets.

We first fix $k = 1$ and evaluate performance as a function of training set size. Performance improves as more training samples are added, confirming the expected behavior of kNN.
We then fix the training set size to 100 positive / 100 negative samples and tune $k$. We find that $k=5$ achieves the best overall performance under this setup (Table~\ref{table:knn-100-100-5}).

\begin{table}[H]
\centering
\begin{tabular}{|c|c|c|c|c|c|c|c|c|}
\hline
\textbf{Experiment} & \multicolumn{4}{c|}{\textbf{Test (Near)}} & \multicolumn{4}{c|}{\textbf{Test (Far)}} \\
\hline
\textbf{Far Group} & IoU & Precision & Recall & F1 & IoU & Precision & Recall & F1 \\
\hline

G0 
& 0.109 & 0.114 & 0.774 & 0.190
& 0.094 & 0.097 & 0.764 & 0.164 \\
\hline

G1 
& 0.096 & 0.099 & 0.807 & 0.169
& 0.121 & 0.128 & 0.787 & 0.208 \\
\hline

G2 
& 0.122 & 0.129 & 0.795 & 0.208
& 0.090 & 0.094 & 0.783 & 0.159 \\
\hline

G3 
& 0.108 & 0.112 & 0.783 & 0.187
& 0.151 & 0.160 & 0.796 & 0.251 \\
\hline

G4 
& 0.115 & 0.120 & 0.815 & 0.196
& 0.094 & 0.096 & 0.821 & 0.165 \\
\hline

\textbf{Avg} 
& \textbf{0.110} & \textbf{0.115} & \textbf{0.795} & \textbf{0.190}
& \textbf{0.110} & \textbf{0.115} & \textbf{0.790} & \textbf{0.189} \\
\hline

\end{tabular}
\caption{KNN on Hedgementation  (k=5, 100 positive/ 100 negative).}\label{table:knn-100-100-5}
\end{table}

To further improve the kNN baseline for hedgerow detection, we incorporate local spatial context by representing each pixel with a concatenation of embeddings from its $3\times3$ neighborhood. The embeddings in this $3\times3$ grid are concatenated in a fixed spatial order, forming a feature vector that captures local structure. Furthermore, to reduce label noise, we only select positive samples that are interior hedgerow pixels: that is, where both the center and neighboring pixels are labeled as positive.
We then add spatial weighting by scaling the center embedding prior to distance computation by a factor. Moderate upweighting improves performance by reducing the influence of noisy neighbors, with a center weight of 6 achieving the best results. Larger weights provide no additional benefit, indicating a trade-off between center emphasis and contextual information.
With this representation, $k=3$ achieves the best performance, while larger values degrade performance due to increased averaging over heterogeneous neighborhoods.
Increasing the number of training samples up to 100{,}000 positive and negative samples, while enforcing per-tile caps to maintain diversity, improves IoU from approximately 0.11 to 0.15 (Table \ref{tab:knn_best_results}). The similar performance on both Near and Far test sets shows that AEF embeddings, trained on the whole world, favor generalization over space compared to training from scratch.

\begin{table}[H]
\centering
\begin{tabular}{|c|c|c|c|c|c|c|c|c|}
\hline
\multicolumn{9}{|c|}{\textbf{KNN on Hedgementation  (k=3, 100,000 positive/ 100,000 negative)}} \\
\hline
\textbf{Experiment} & \multicolumn{4}{c|}{\textbf{Test (Near)}} & \multicolumn{4}{c|}{\textbf{Test (Far)}} \\
\hline
\textbf{Far Group} & IoU & Precision & Recall & F1 & IoU & Precision & Recall & F1 \\
\hline

G0 
& 0.155 & 0.166 & 0.735 & 0.257
& 0.134 & 0.141 & 0.729 & 0.222 \\
\hline

G1 
& 0.147 & 0.156 & 0.738 & 0.244
& 0.158 & 0.170 & 0.728 & 0.260 \\
\hline

G2 
& 0.155 & 0.166 & 0.740 & 0.256
& 0.119 & 0.125 & 0.728 & 0.205 \\
\hline

G3 
& 0.145 & 0.153 & 0.736 & 0.255
& 0.173 & 0.187 & 0.735 & 0.230 \\
\hline

G4 
& 0.155 & 0.164 & 0.751 & 0.196
& 0.136 & 0.144 & 0.749 & 0.165 \\
\hline

\textbf{Avg} 
& \textbf{0.151} & \textbf{0.161} & \textbf{0.740} & \textbf{0.251}
& \textbf{0.147} & \textbf{0.157} & \textbf{0.733} & \textbf{0.245} \\
\hline

\end{tabular}
\caption{Performance of the best kNN configuration on the Near and Far test sets. Results are reported across Far Groups (G0–G4), with averages shown in the last row. The model uses $3\times3$ neighborhood concatenation, center weighting of 6, $k=3$, and 100{,}000 positive/100{,}000 negative training samples.}\label{tab:knn_best_results}
\end{table}

%% file: text/appendix/results_rf_space.tex
We also fit a Random Forest model on the usefulness of AEF embeddings \cite{aef} for the hedgerow detection task. We tune the following hyper-parameters: number of estimators ($N_{est}$), minimum samples per leaf ($N_{min\_leaf\_samples}$), maximum depth of each tree ($N_{max\_tree\_depth}$), and the probability threshold for prediction ($proba\_threshold)$.

We also set class weights to balanced, to counteract the imbalance of our dataset (overwhelming number of negatives compared to positives). Balancing class weights assigns a higher weight to the minority class (positives) and a lower weight to the majority class (negatives) when fitting the model. Without balanced weights, our model just predicts negative for all points to achieve great accuracy, compromising IoU.

We use 100{,}000 positive and negative samples to compare with the kNN, and the whole training set to maximize performance.
We tune parameters sequentially. We first fix the probability threshold at $0.5$ and search over the following space: $N_{est}$ between 100 and 1000, $N_{min\_leaf\_samples}$ between 1 and 5, $N_{max\_tree\_depth}$ in [$20,40,\infty$], and pick the combination with highest validation IoU:
$N_{est} = 500$,  $N_{min\_leaf\_samples} = 2$,  $N_{max\_tree\_depth} = \infty$.
(We note that $N_{est} = 1000$ achieves a slightly better IoU than $N_{est} = 500$, around 0.007\% higher, but the training time is 50\% longer for such a minor gain).

We then tune the probability threshold ($proba\_threshold$) between 0.1 and 0.9. The best $proba\_threshold$ is different for the two training set variations, because one uses a down-sampled balanced dataset while the other uses the entire unbalanced training set. We conduct the space-extrapolation experiments using the best probability thresholds for the two variations respectively, with $proba\_threshold = 0.7$ for the balanced 100{,}000 positive / 100{,}000 negative samples model, and $proba\_threshold = 0.2$ for the one that uses the entire training set.

\begin{table}[H]
\centering
\begin{tabular}{|c|c|c|c|c|c|c|c|c|}
\hline
\multicolumn{9}{|c|}{\textbf{Random Forest on Hedgementation (100{,}000 Pos / Neg Samples)}} \\
\hline
\multicolumn{9}{|c|}{N$_{est}$ = 500,  N$_{min\_leaf\_samples}$ = 2,  N$_{max\_tree\_depth}$ = $\infty$, proba\_threshold=0.7} \\
\hline
\textbf{Experiment} & \multicolumn{4}{c|}{\textbf{Test (Near)}} & \multicolumn{4}{c|}{\textbf{Test (Far)}} \\
\hline
\textbf{Far Group} & IoU & Precision & Recall & F1 & IoU & Precision & Recall & F1 \\
\hline

G0 
& 0.226 & 0.324 & 0.501 & 0.389
& 0.213 & 0.282 & 0.526 & 0.373 \\
\hline

G1 
& 0.221 & 0.323 & 0.492 & 0.388
& 0.233 & 0.308 & 0.558 & 0.382 \\
\hline

G2 
& 0.231 & 0.317 & 0.533 & 0.394
& 0.205 & 0.312 & 0.443 & 0.377 \\
\hline

G3 
& 0.218 & 0.311 & 0.506 & 0.384
& 0.239 & 0.346 & 0.500 & 0.387 \\
\hline

G4 
& 0.221 & 0.312 & 0.499 & 0.379
& 0.236 & 0.360 & 0.538 & 0.433 \\
\hline

\textbf{Avg} 
& \textbf{0.223} & \textbf{0.317} & \textbf{0.506} & \textbf{0.387}
& \textbf{0.225} & \textbf{0.321} & \textbf{0.513} & \textbf{0.391} \\
\hline

\end{tabular}
\caption{Performance of the Random Forest configuration on the Near and Far test sets using balanced 100{,}000 positive/negative samples. Results are reported across Far Groups (G0–G4), with averages shown in the last row. The model uses 500 trees, minimum samples per leaf as 2, infinite maximum tree depth, probability threshold as 0.7, and 100,000 positive / 100,000 negative samples.}
\label{tab:rf_100000_results}
\end{table}

Table \ref{tab:rf_100000_results} shows results with 100{,}000 positive and 100{,}000 negative training samples, which is directly comparable to the kNN.
The random forest classifier produces much better IoU than the best kNN with the same number of samples, outperforming it on average by 7\% points of IoU. It also has a 16\% increase on precision (but sees a 20\% drop in recall) leading to an overall better F1. 
As with the kNN, the performance is almost identical for near and far test sets across groups, showing that the AEF embeddings enable good generalization over space.

\begin{table}[H]
\centering
\begin{tabular}{|c|c|c|c|c|c|c|c|c|}
\hline
\multicolumn{9}{|c|}{\textbf{Random Forest on Hedgementation}} \\
\hline
\multicolumn{9}{|c|}{N$_{est}$ = 500,  N$_{min\_leaf\_samples}$ = 2,  N$_{max\_tree\_depth}$ = $\infty$, proba\_threshold=0.2} \\
\hline
\textbf{Experiment} & \multicolumn{4}{c|}{\textbf{Test (Near)}} & \multicolumn{4}{c|}{\textbf{Test (Far)}} \\
\hline
\textbf{Far Group} & IoU & Precision & Recall & F1 & IoU & Precision & Recall & F1 \\
\hline

G0 
& 0.236 & 0.358 & 0.471 & 0.407
& 0.228 & 0.328 & 0.488 & 0.397 \\
\hline

G1 
& 0.232 & 0.356 & 0.463 & 0.407
& 0.245 & 0.346 & 0.513 & 0.400 \\
\hline

G2 
& 0.243 & 0.356 & 0.497 & 0.412
& 0.212 & 0.348 & 0.417 & 0.387 \\
\hline

G3 
& 0.231 & 0.350 & 0.473 & 0.406
& 0.249 & 0.369 & 0.475 & 0.404 \\
\hline

G4 
& 0.232 & 0.349 & 0.467 & 0.396
& 0.249 & 0.380 & 0.509 & 0.456 \\
\hline

\textbf{Avg} 
& \textbf{0.235} & \textbf{0.354} & \textbf{0.474} & \textbf{0.406}
& \textbf{0.237} & \textbf{0.354} & \textbf{0.480} & \textbf{0.409} \\
\hline

\end{tabular}
\caption{Performance of the best Random Forest configuration on the Near and Far test sets. Results are reported across Far Groups (G0–G4), with averages shown in the last row. The model uses 500 trees, minimum samples per leaf as 2, infinite maximum tree depth, probability threshold as 0.2, and the entire training set.}
\label{tab:tf_best_results}
\end{table}
\FloatBarrier

Table \ref{tab:tf_best_results} shows results using the full training set, yielding a 1.2\% point improvement in IoU compared to a $200k$ balanced training set.
The performance reaches an IoU of 0.237 (on test Far), a significant improvement over the kNN, but still far from deep learning models trained from scratch: our best model, the PASTIS U-TAE reaches 38.8\% IoU on test Far, and 41\% on test Near.
However, compared to the PASTIS U-TAE model, the Random Forest model on AEF embeddings fully generalizes over space: the Near and Far IoUs are almost identical (a difference of 0.2\% point change in IoU, slightly better on the Far test), compared to a 2.1\% point drop on the Far test for the PASTIS model.

%% file: text/appendix/results_logistic_regression_space.tex
Finally, we fit a Logistic Regression model using the AEF \cite{aef} embeddings. We use an $\ell_2$ regularization using the \textit{saga} solver with regularization parameter $C=1.0$. The training data is constructed by sampling up to 1000 pixels per tile, followed by global balancing between hedgerow and non-hedgerow classes.

We perform hyper-parameter tuning over the prediction threshold $\tau \in \{0.5, 0.6, 0.7, 0.8, 0.9\}$. We observe that performance improves as $\tau$ increases up to $\tau=0.8$, and degrades beyond this point.
Table~\ref{tab:logreg_results} shows that at $\tau=0.8$, the model reaches an IoU of 20\% on both test Near and Far, close to (but lower than) the Random Forest. Once again, using the AEF embedding shows strong generalization over space (a 0.6\% drop of IoU between Near and Far tests, against 2.1\% for PASTIS' U-TAE).

\begin{table}[H]
\centering
\begin{tabular}{|c|c|c|c|c|c|c|c|c|}
\hline
\multicolumn{9}{|c|}{\textbf{Logistic Regression on Hedgementation }} \\
\hline
\textbf{Experiment} & \multicolumn{4}{c|}{\textbf{Test (Near)}} & \multicolumn{4}{c|}{\textbf{Test (Far)}} \\
\hline
\textbf{Far Group} & IoU & Precision & Recall & F1 & IoU & Precision & Recall & F1 \\
\hline

G0 
& 0.206 & 0.297 & 0.446 & 0.325
& 0.198 & 0.253 & 0.495 & 0.310 \\
\hline

G1 
& 0.202 & 0.279 & 0.453 & 0.318
& 0.214 & 0.309 & 0.499 & 0.337 \\
\hline

G2 
& 0.211 & 0.296 & 0.465 & 0.330
& 0.172 & 0.253 & 0.371 & 0.277 \\
\hline

G3 
& 0.200 & 0.283 & 0.451 & 0.317
& 0.217 & 0.340 & 0.407 & 0.335 \\
\hline

G4 
& 0.209 & 0.292 & 0.473 & 0.327
& 0.193 & 0.257 & 0.485 & 0.309 \\
\hline

\textbf{Avg} 
& \textbf{0.206} & \textbf{0.289} & \textbf{0.457} & \textbf{0.323}
& \textbf{0.200} & \textbf{0.283} & \textbf{0.450} & \textbf{0.314} \\
\hline

\end{tabular}
\caption{Performance of logistic regression on the Near and Far test sets across Far Groups (G0–G4).}
\label{tab:logreg_results}
\end{table}
\FloatBarrier

%% file: text/appendix/results_pastis_climate-zones.tex
Table \ref{tab:pastis_climate_zones} shows the results of the PASTIS U-TAE model on the Hedgementation task evaluating generalizations over agriculturally-relevant climate zones.
We make three observations.

First, the size of the training set has a large impact on model performance, with a 4.8\% point drop in IoU on a uniformly subsampled dataset the size of the subtropic dataset. We thus subsample all datasets to the smallest size to make performance comparable.

Second, subtropic patches are consistently the most difficult to segment, even for models trained exclusively on those patches. As one may expect, the difference is at it's smallest for the subtropic-only model, where it is only 2.1\%, and at it's largest for the temperate-only trained model, where it is 7.0\%.

Third, while the downsampled model performs worse on subtropic tiles than the subtropic-only model, it performs better on temperate tiles than the temperate-only model.
It also performs best on the full test set.
This may indicate that subtropic tiles contain important information useful to generalize to subtropic tiles, while a diversity of climate zones is important for genralization.

\begin{table}[H]
\centering
\begin{tabular}{|l|l|l|c|c|c|c|}
\hline
\textbf{Architecture} & \textbf{Trained On} & \textbf{Tested On} & \textbf{IoU} & \textbf{Precision} & \textbf{Recall} & \textbf{F1} \\
\hline
PASTIS & Full Dataset & Full Dataset & 0.416 & 0.538 & 0.646 & 0.587 \\
\hline
PASTIS & Full Dataset & Temp. Only & 0.424 & 0.542 & 0.661 & 0.596 \\
\hline
PASTIS & Full Dataset & Subtropic Only & 0.374 & 0.518 & 0.574 & 0.544 \\
\hline
\hline
PASTIS & Subtropic Only & Full Dataset & 0.352 & 0.453 & 0.613 & 0.521 \\
\hline
PASTIS & Subtropic Only & Temp. Only & 0.357 & 0.46 & 0.615 & 0.526 \\
\hline
PASTIS & Subtropic Only & Subtropic Only & 0.331 & 0.422 & 0.604 & 0.497 \\
\hline
\hline
PASTIS & Temperate Only & Full Dataset & 0.357 & 0.456 & 0.621 & 0.526 \\
\hline
PASTIS & Temperate Only & Temp. Only & 0.372 & 0.47 & 0.642 & 0.542 \\
\hline
PASTIS & Temperate Only & Subtropic Only & 0.287 & 0.392 & 0.52 & 0.447 \\
\hline
\hline
PASTIS & Subs. Dataset & Full Dataset & 0.368 & 0.468 & 0.634 & 0.538 \\
\hline
PASTIS & Subs. Dataset & Temp. Only & 0.38 & 0.478 & 0.648 & 0.55 \\
\hline
PASTIS & Subs. Dataset & Subtropic Only & 0.316 & 0.419 & 0.563 & 0.48 \\
\hline
\end{tabular}

\caption{Results across climate zones for models trained all, only subtropic, and only temperate patches.  All datasets after the first group are random subsets matching the size of the subtropic only dataset (the smallest subset).}
\label{tab:pastis_climate_zones}
\end{table}

%% file: text/appendix/results_ftw_climate-zones.tex
Table~\ref{table:ftw_thz} shows the impact of climate zones on the FTW model.
Again, subtropic patches are consistently the most difficult across all models, even for the model trained exclusively on subtropic patches. The gap between subtropic and temperate performance varies by training set. As expected, this difference is smallest for the subtropic-only model, where it is only 2.3\% (0.184 vs. 0.161 IoU), and largest for the temperate-only model, where it reaches 2.1\% (0.1234 vs. 0.102 IoU).

Interestingly, the subsampled (but complete) training set outperforms significantly outperforms all other models. For instance, it outperforms the subtropic-only model even when testing on subtropic tiles (0.271 vs. 0.161 IoU) and also outperforms the temperate-only model on temperate tiles (0.331 vs. 0.1234 IoU). This suggests that exposure to diverse geographic contexts during training provides generalization benefits that outweigh the reduction in per-class sample count.

\begin{table}[H]
\centering
\begin{tabular}{|l|l|l|c|c|c|c|}
\hline
\textbf{Architecture} & \textbf{Trained On} & \textbf{Tested On} & \textbf{IoU} & \textbf{Precision} & \textbf{Recall} & \textbf{F1} \\
\hline
FTW & Full Dataset & Full Dataset & 0.333 & 0.6069 & 0.4246 & 0.4996 \\
\hline
FTW & Full Dataset & Temp. Only & 0.3432 & 0.616 & 0.4365 & 0.511 \\
\hline
FTW & Full Dataset & Subtropic Only & 0.2839 & 0.5583 & 0.3661 & 0.4422 \\
\hline
\hline
FTW & Subtropic Only & Full Dataset & 0.1799 & 0.6061 & 0.2037 & 0.3049 \\
\hline
FTW & Subtropic Only & Temp. Only & 0.184 & 0.6194 & 0.2074 & 0.3107 \\
\hline
FTW & Subtropic Only & Subtropic Only & 0.161 & 0.5463 & 0.1859 & 0.2774 \\
\hline
\hline
FTW & Temperate Only & Full Dataset & 0.1197 & 0.6994 & 0.1262 & 0.2138 \\
\hline
FTW & Temperate Only & Temp. Only & 0.1234 & 0.7174 & 0.1297 & 0.2197 \\
\hline
FTW & Temperate Only & Subtropic Only & 0.102 & 0.6104 & 0.1091 & 0.1852 \\
\hline
\hline
FTW & Subs. Dataset & Full Dataset & 0.3208 & 0.6237 & 0.3978 & 0.4857 \\
\hline
FTW & Subs. Dataset & Temp. Only & 0.331 & 0.632 & 0.41 & 0.4974 \\
\hline
FTW & Subs. Dataset & Subtropic Only & 0.271 & 0.5782 & 0.3378 & 0.4264 \\
\hline
\end{tabular}
\caption{FTW results by THZ Class on Hedgementation. All datasets after the first group are random subsets matching the size of the subtropic only dataset (the smallest subset).}\label{table:ftw_thz}
\end{table}

%% file: text/appendix/ag_exps.tex
To distinguish agricultural areas from non-agricultural areas, we use data from the \emph{Registre Parcellaire Graphique} (RPG), which contains spatial polygons demarcating agricultural parcels across more than 90\% of the agricultural zone \citep{cantelaube2014registre}. By overlapping our patches with these polygons and rasterizing the result into a binary mask, we can assign every pixel in our dataset as either agricultural or non-agricultural. This information is relevant for multiple reasons.
First, we expect the BD Haies labels to have higher quality in agricultural zones, where they are validated by individual farmers. In addition, hedgerows are prevalent in agricultural areas, and particularly important to monitor there. This lets us focus on our models' error in agricultural zones explicitly, and measure any distribution differences in hedgerow labels and their detection through satellite images in agricultural vs. non agricultural landscapes.

Since many hedgerows are found adjacent to agricultural areas rather than directly within them, we also experiment with buffering the border of all agricultural polygons by a constant 10 meters before rasterizing, in order to include such hedgerows.

\begin{table}[t]
\centering
\begin{tabular}{|l|l|c|l|c|}
\hline
\textbf{Pixels} & \textbf{Total Pixels} & \textbf{\% of Total} & \textbf{Hedgerow Pixels } & \textbf{\% of Hedgerow} \\
\hline
All & 49,070,080 & 100\% & 2,430,758 & 100\% \\
\hline
Agri. + 10M Buffer & 32,737,848 & 67\% & 2,010,352 & 83\% \\
\hline
Agricultural & 31,143,086 & 63\% & 1,751,315 & 71\% \\
\hline
Non-Agri. & 17,926,994 & 36\% & 679,443 & 28\% \\
\hline
Non-Agri. - 10M Buffer & 16,332,232 & 33\% & 420,406 & 17\% \\
\hline

\end{tabular}
\caption{Tables showing the proportion of all pixels, and hedgerow pixels specifically, included by agricultural masks.}
    \label{fig:pixel_ag_dist}
\end{table}

\begin{figure}
    \centering
    \includegraphics[width=0.5\linewidth]{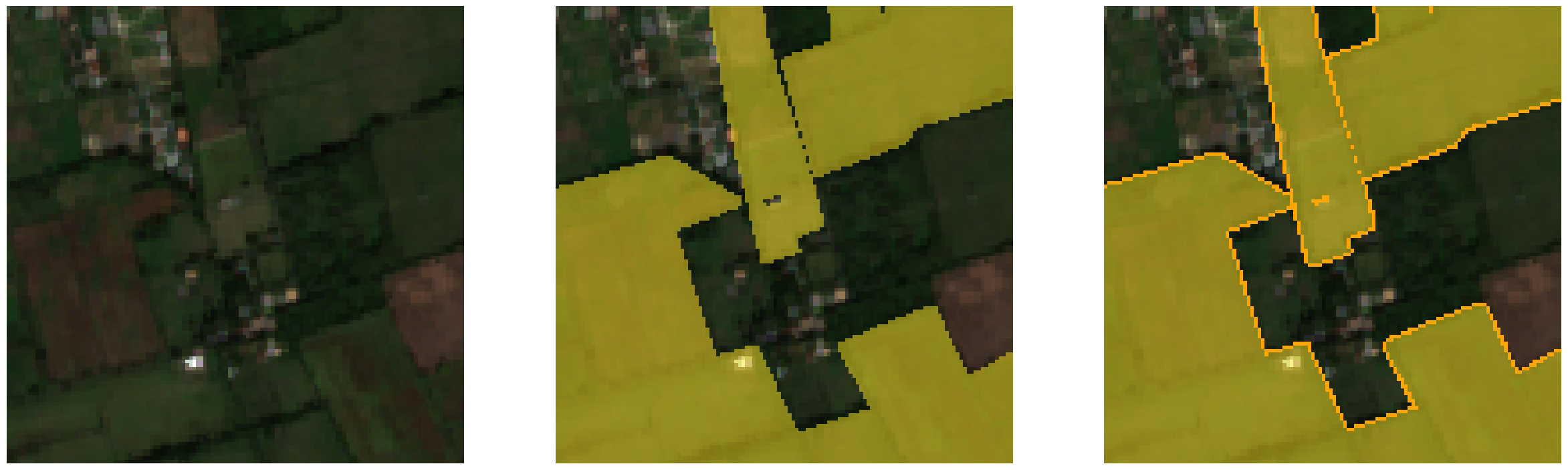}
    \caption{The construction of the RPG agriculture mask. The mask is shown in yellow, and the 10-m buffer is shown in orange.}
    \label{fig:rpg_example}
\end{figure}

We see that 63\%, a significant majority, of pixels in our dataset are classified as falling within agricultural areas when referencing the RPG data. The proportion is even larger when we consider specifically hedgerow pixels, of which 71\% are found in agricultural areas. Hedgerow pixels are also very likely to be situated directly on the border of agricultural areas: adding the 10 meter buffer increases overall pixel coverage by 4\%(from 63\% to 67\%), but hedgerow coverage by 12\% (from 71\% to 83\%).

To assess variation in hedgerow segmentation performance in agricultural vs. non-agricultural landscapes, we evaluate our models on these subsets of the test set (with and without buffers), and compare the metrics.
We also assess the importance of agricultural vs. non-agricultural pixels to training by training models exclusively on those subsets (in this case, we mask the loss of excluded pixels by multiplying it by zero), and comparing their results.
This however introduces a confounder as it varies the size of the training set, which we observe the have a significant impact on performance (Appendix \ref{appendix:space}).
We also train models by randomly masking out the loss from pixels until the number of pixels used is the same as the agriculture-only experiment. This way, all models ``see'' the same number of pixels during training, and we can better isolate the effects to the distribution of training pixels.

%% file: text/appendix/results_pastis_ag.tex
Table \ref{tab:pastis_train_ag_pixels} shows the performance of models trained on a mix of all pixels or just agricultural pixels, on all subsets of pixels. Table \ref{tab:pastis_train_ag_pixels} focuses on training on non-ag pixels. We make two main observations.

First, training on all pixels always yields better results, though the effect is small compared to ag only pixels. Since the size of dataset matters, it seems like the best strategy is to train on all available data.

Second, hedgerows in agricultural landscapes are significantly easier to map: the IoU of the PASTIS U-TAE model reaches 45.9\% of IoU on agricultural pixels, up from 41\% on all pixels.
This is a valuable and encouraging findings for applications that focus on these landscapes.

\begin{table}[t]
\centering
\begin{tabular}{|l|l|l|c|c|c|c|}
\hline
\textbf{Model} & \textbf{Trained On} & \textbf{Tested On} & \textbf{IoU} & \textbf{Presc.} & \textbf{Recall} & \textbf{F1} \\
\hline
PASTIS & All Pixels & All Pixels & 0.416 & 0.538 & 0.646 & 0.587 \\
\hline
PASTIS & All Pixels (subs.) & All Pixels &  0.41  & 0.507&  0.68 &  0.581 \\
\hline
PASTIS & Agri. Pixels & All Pixels & 0.378 & 0.448 & 0.706 & 0.548 \\
\hline
PASTIS & Agri.  + Buffer  & All Pixels &  0.394 & 0.492 & 0.665 & 0.566 \\
\hline
\hline

\hline
PASTIS & All Pixels & Agri. Pixels & 0.462 & 0.576 & 0.7 & 0.632 \\
\hline
PASTIS & All Pixels (subs.) & Agri. Pixels & 0.459 & 0.555 & 0.726 & 0.629 \\
\hline
PASTIS & Agri. Pixels & Agri. Pixels & 0.457 & 0.545 & 0.739 & 0.627 \\
\hline
PASTIS & Agri.  + Buffer & Agri. Pixels & 0.452 & 0.555 & 0.709 & 0.622 \\
\hline
\hline

\hline
PASTIS & All Pixels & Agri.  + Buffer & 0.457 & 0.572 & 0.694 & 0.627 \\
\hline
PASTIS & All Pixels (subs.) & Agri.  + Buffer & 0.454 & 0.55 &  0.722 & 0.624 \\
\hline
PASTIS & Agri. Pixels & Agri.  + Buffer & 0.451 & 0.538 & 0.737 & 0.622 \\
\hline
PASTIS & Agri.  + Buffer & Agri.  + Buffer & 0.447 & 0.551 & 0.704 & 0.618 \\
\hline

\hline
PASTIS & All Pixels & Non-Agri. Pixels & 0.303 & 0.433 & 0.502 & 0.465 \\
\hline
PASTIS & All Pixels (subs.) & Non-Agri. Pixels & 0.297 & 0.39 &  0.556 & 0.458 \\
\hline
PASTIS & Agri. Pixels & Non-Agri. Pixels & 0.243 & 0.286 & 0.617 & 0.391 \\
\hline
PASTIS & Agri.  + Buffer & Non-Agri. Pixels & 0.273 & 0.354 & 0.546 & 0.429 \\
\hline
\hline

\hline
PASTIS & All Pixels (subs.) &  Non-Agri. - Buffer & 0.234 & 0.356 & 0.406 & 0.379 \\
\hline
PASTIS & All Pixels (subs.) &  Non-Agri. - Buffer & 0.232 & 0.316 & 0.466 & 0.376 \\
\hline
PASTIS & Agri. Pixels & Non-Agri. - Buffer & 0.179 & 0.21 & 0.549 & 0.304 \\
\hline
PASTIS & Agri.  + Buffer & Non-Agri. - Buffer & 0.206 & 0.27 &  0.465 & 0.342  \\
\hline
\end{tabular}
\caption{Results across different pixel sets for models trained on all pixels, Agri. pixels, and Agri. pixels + Buffer.}
\label{tab:pastis_train_ag_pixels}
\end{table}

\begin{table}[t]
\centering
\begin{tabular}{|l|l|l|c|c|c|c|}
\hline
\textbf{Model} & \textbf{Trained On} & \textbf{Tested On} & \textbf{IoU} & \textbf{Presc.} & \textbf{Recall} & \textbf{F1} \\
\hline
PASTIS & All Pixels(subs.) & All Pixels &  0.39 &  0.483 & 0.67  & 0.561 \\
\hline
PASTIS & Non-Agri. Pixels & All Pixels &  0.205 & 0.232 &  0.641 & 0.34 \\
\hline
PASTIS & Non-Agri.  - Buffer  & All Pixels &  0.242 & 0.3 & 0.555 & 0.39 \\
\hline
\hline

\hline
PASTIS & All Pixels (subs.) & Agri. Pixels & 0.435 & 0.527 & 0.713 & 0.606 \\
\hline
PASTIS & Non-Agri. Pixels & Agri. Pixels & 0.191 & 0.209 & 0.689 & 0.321 \\
\hline
PASTIS & Non-Agri.  - Buffer & Agri. Pixels & 0.236 & 0.281 & 0.593 & 0.382 \\
\hline
\hline

\hline
PASTIS & All Pixels  (subs.) & Agri. + Buffer & 0.431 & 0.524 & 0.709 & 0.603 \\
\hline
PASTIS & Non-Agri. Pixels & Agri.  + Buffer & 0.204 & 0.225 & 0.683  & 0.339 \\
\hline
PASTIS & Non-Agri.  - Buffer & Agri.  + Buffer & 0.204 & 0.225 & 0.683 & 0.339 \\
\hline
\hline

PASTIS & All Pixels (subs.) & Non-Agri. Pixels & 0.287 & 0.374 & 0.553 & 0.446 \\
\hline
PASTIS & Non-Agri. Pixels & Non-Agri. Pixels & 0.278 & 0.379 & 0.511  &0.435 \\
\hline
PASTIS & Non-Agri.  - Buffer & Non-Agri. Pixels &  0.267 & 0.393  & 0.453 & 0.421 \\
\hline
\hline

\hline
PASTIS & All Pixels (subs.) &  Non-Agri. - Buffer & 0.225 & 0.301 & 0.472 & 0.368 \\
\hline
PASTIS & Non-Agri. Pixels & Non-Agri. - Buffer & 0.215 & 0.303 &  0.425 & 0.354 \\
\hline
PASTIS & Non-Agri. - Buffer & Non-Agri. - Buffer & 0.21  & 0.314 & 0.386 & 0.347  \\
\hline
\end{tabular}

\caption{Results across different pixel sets for models trained on all pixels, Non-Agri. pixels, and Non-Agri. pixels - Buffer.}
\label{tab:pastis_train_ag_pixels}
\end{table}

%% file: text/appendix/results_ftw_ag.tex
A similar experiment with the FTW model yields qualitatively consistent findings, as shown on Table~\ref{table:ftw-results-agri}. All IoUs are lower though, as the model is less performant that the PASTIS U-TAE on this task.

\begin{table}[t]
\centering
\begin{tabular}{|l|l|l|c|c|c|c|}
\hline
\textbf{Model} & \textbf{Trained On} & \textbf{Tested On} & \textbf{IoU} & \textbf{Prec} & \textbf{R} & \textbf{F1} \\
\hline
FTW & All Pixels & All Pixels & 0.333 & 0.607 & 0.425 & 0.500 \\
\hline
FTW & Agri. + Buffer & All Pixels & 0.310 & 0.608 & 0.388 & 0.474 \\
\hline
FTW & Agri. - Buffer & All Pixels & 0.326 & 0.601 & 0.415 & 0.491 \\
\hline
FTW & Non-Agri. + Buffer & All Pixels & 0.126 & 0.728 & 0.132 & 0.224 \\
\hline
FTW & Non-Agri. - Buffer & All Pixels & 0.201 & 0.648 & 0.226 & 0.335 \\
\hline\hline
FTW & All Pixels & Agri. + Buffer & 0.367 & 0.639 & 0.462 & 0.537 \\
\hline
FTW & Agri. + Buffer & Agri. + Buffer & 0.345 & 0.658 & 0.421 & 0.513 \\
\hline
FTW & Agri. - Buffer & Agri. + Buffer & 0.360 & 0.655 & 0.444 & 0.529 \\
\hline
FTW & Non-Agri. + Buffer & Agri. + Buffer & 0.140 & 0.752 & 0.146 & 0.245 \\
\hline
FTW & Non-Agri. - Buffer & Agri. + Buffer & 0.222 & 0.671 & 0.249 & 0.363 \\
\hline\hline
FTW & All Pixels & Non-Agri. + Buffer & 0.173 & 0.403 & 0.233 & 0.296 \\
\hline
FTW & Agri. + Buffer & Non-Agri. + Buffer & 0.156 & 0.350 & 0.221 & 0.271 \\
\hline
FTW & Agri. - Buffer & Non-Agri. + Buffer & 0.180 & 0.356 & 0.268 & 0.306 \\
\hline
FTW & Non-Agri. + Buffer & Non-Agri. + Buffer & 0.059 & 0.528 & 0.062 & 0.111 \\
\hline
FTW & Non-Agri. - Buffer & Non-Agri. + Buffer & 0.097 & 0.465 & 0.110 & 0.177 \\
\hline\hline
FTW & All Pixels & Agri. - Buffer & 0.373 & 0.648 & 0.468 & 0.544 \\
\hline
FTW & Agri. + Buffer & Agri. - Buffer & 0.352 & 0.668 & 0.426 & 0.521 \\
\hline
FTW & Agri. - Buffer & Agri. - Buffer & 0.367 & 0.669 & 0.448 & 0.537 \\
\hline
FTW & Non-Agri. + Buffer & Agri. - Buffer & 0.143 & 0.760 & 0.150 & 0.250 \\
\hline
FTW & Non-Agri. - Buffer & Agri. - Buffer & 0.224 & 0.675 & 0.251 & 0.366 \\
\hline\hline
FTW & All Pixels & Non-Agri. - Buffer & 0.231 & 0.482 & 0.307 & 0.375 \\
\hline
FTW & Agri. + Buffer & Non-Agri. - Buffer & 0.210 & 0.444 & 0.284 & 0.346 \\
\hline
FTW & Agri. - Buffer & Non-Agri. - Buffer & 0.229 & 0.437 & 0.325 & 0.373 \\
\hline
FTW & Non-Agri. + Buffer & Non-Agri. - Buffer & 0.081 & 0.609 & 0.086 & 0.150 \\
\hline
FTW & Non-Agri. - Buffer & Non-Agri. - Buffer & 0.140 & 0.553 & 0.158 & 0.246 \\
\hline
\end{tabular}
\caption{FTW model results across different training and testing pixel masks.}
\label{table:ftw-results-agri}
\end{table}

%% file: iclr2026_conference.bib
@article{biffi2023planting,
  title={Planting hedgerows: Biomass carbon sequestration and contribution towards net-zero targets},
  author={Biffi, Sofia and Chapman, Pippa J and Grayson, Richard P and Ziv, Guy},
  journal={Science of the Total Environment},
  volume={892},
  pages={164482},
  year={2023},
  publisher={Elsevier}
}

@article{asbjornsen2014targeting,
  title={Targeting perennial vegetation in agricultural landscapes for enhancing ecosystem services},
  author={Asbjornsen, Heidi and Hernandez-Santana, Virginia and Liebman, Matthew and Bayala, Jules and Chen, Jiquan and Helmers, Matthew and Ong, CK and Schulte, LA},
  journal={Renewable Agriculture and Food Systems},
  volume={29},
  number={2},
  pages={101--125},
  year={2014},
  publisher={Cambridge University Press}
}

@article{montgomery2020hedgerows,
  title={Hedgerows as ecosystems: service delivery, management, and restoration},
  author={Montgomery, Ian and Caruso, Tancredi and Reid, Neil},
  journal={Annual Review of Ecology, Evolution, and Systematics},
  volume={51},
  number={1},
  pages={81--102},
  year={2020},
  publisher={Annual Reviews}
}

@article{albrecht2020effectiveness,
  title={The effectiveness of flower strips and hedgerows on pest control, pollination services and crop yield: a quantitative synthesis},
  author={Albrecht, Matthias and Kleijn, David and Williams, Neal M and Tschumi, Matthias and Blaauw, Brett R and Bommarco, Riccardo and Campbell, Alistair J and Dainese, Matteo and Drummond, Francis A and Entling, Martin H and others},
  journal={Ecology letters},
  volume={23},
  number={10},
  pages={1488--1498},
  year={2020},
  publisher={Wiley Online Library}
}

@article{muro2025hedgerow,
  title={Hedgerow mapping with high resolution satellite imagery to support policy initiatives at national level},
  author={Muro, Javier and Blickensd{\"o}rfer, Lukas and Don, Axel and K{\"o}ber, Anna and Asam, Sarah and Schwieder, Marcel and Erasmi, Stefan},
  journal={Remote Sensing of Environment},
  volume={328},
  pages={114870},
  year={2025},
  publisher={Elsevier}
}

@inproceedings{faucqueur2019new,
  title={A new Copernicus high resolution layer at pan-European scale: Small woody features},
  author={Faucqueur, Loic and Morin, Nathalie and Masse, Antoine and Remy, Pierre-Yves and Hug{\'e}, Justine and Kenner, Cl{\'e}mence and Dazin, Fabrice and Descl{\'e}e, Baudouin and Sannier, Christophe},
  booktitle={Remote Sensing for Agriculture, Ecosystems, and Hydrology XXI},
  volume={11149},
  pages={268--278},
  year={2019},
  organization={SPIE}
}

@article{ha2019shelterbelt,
  title={Shelterbelt agroforestry systems inventory and removal analyzed by object-based classification of satellite data in Saskatchewan, Canada},
  author={Ha, Thuan V and Amichev, Beyhan Y and Belcher, Kenneth W and Bentham, Murray J and Kulshreshtha, Suren N and Laroque, Colin P and Van Rees, Ken CJ},
  journal={Canadian Journal of Remote Sensing},
  volume={45},
  number={2},
  pages={246--263},
  year={2019},
  publisher={Taylor \& Francis}
}

@article{liu2023overlooked,
  title={The overlooked contribution of trees outside forests to tree cover and woody biomass across Europe},
  author={Liu, Siyu and Brandt, Martin and Nord-Larsen, Thomas and Chave, Jerome and Reiner, Florian and Lang, Nico and Tong, Xiaoye and Ciais, Philippe and Igel, Christian and Pascual, Adrian and others},
  journal={Science Advances},
  volume={9},
  number={37},
  pages={eadh4097},
  year={2023},
  publisher={American Association for the Advancement of Science}
}

@misc{BDhaie2024,
  author={IGN, Institut national de l'information g\'eographique et foresti\`ere},
  title={BD Haie V2 mars 2024},
  year={2024},
  note={Accessed: 2025-08-01 from {Link: https://geoservices.ign.fr/bdhaie}}
}

@misc{sentinel2harmonized,
  author={ESA, European Space Agency and Copernicus and Google Earth Engine},
  title={Harmonized Sentinel-2 MSI: MultiSpectral Instrument, Level-2A (SR)},
  year={2025},
  note={Accessed: 2025-08-01 from {https://developers.google.com/earth-engine/datasets/catalog/COPERNICUS\_S2\_SR\_HARMONIZED}}
}

@article{drexler2024carbon,
  title={Carbon sequestration potential in hedgerow soils: Results from 23 sites in Germany},
  author={Drexler, Sophie and Don, Axel},
  journal={Geoderma},
  volume={445},
  pages={116878},
  year={2024},
  publisher={Elsevier}
}

@article{gorelick2017google,
title={Google Earth Engine: Planetary-scale geospatial analysis for everyone},
author={Gorelick, Noel and Hancher, Matt and Dixon, Mike and Ilyushchenko, Simon and Thau, David and Moore, Rebecca},
journal={Remote Sensing of Environment},
year={2017},
publisher={Elsevier},
doi={10.1016/j.rse.2017.06.031},
url={https://doi.org/10.1016/j.rse.2017.06.031}
}

@misc{pastis,
Author = {Vivien Sainte Fare Garnot and Loic Landrieu},
Title = {Panoptic Segmentation of Satellite Image Time Series with Convolutional Temporal Attention Networks},
Year = {2021},
Eprint = {arXiv:2107.07933},
}

@misc{aef,
Author = {Christopher F. Brown and Michal R. Kazmierski and Valerie J. Pasquarella and William J. Rucklidge and Masha Samsikova and Chenhui Zhang and Evan Shelhamer and Estefania Lahera and Olivia Wiles and Simon Ilyushchenko and Noel Gorelick and Lihui Lydia Zhang and Sophia Alj and Emily Schechter and Sean Askay and Oliver Guinan and Rebecca Moore and Alexis Boukouvalas and Pushmeet Kohli},
Title = {AlphaEarth Foundations: An embedding field model for accurate and efficient global mapping from sparse label data},
Year = {2025},
Eprint = {arXiv:2507.22291},
}

@misc{ftw,
Author = {Hannah Kerner and Snehal Chaudhari and Aninda Ghosh and Caleb Robinson and Adeel Ahmad and Eddie Choi and Nathan Jacobs and Chris Holmes and Matthias Mohr and Rahul Dodhia and Juan M. Lavista Ferres and Jennifer Marcus},
Title = {Fields of The World: A Machine Learning Benchmark Dataset For Global Agricultural Field Boundary Segmentation},
Year = {2024},
Eprint = {arXiv:2409.16252},
}

@misc{fao_gaez_v4,
  author = {{FAO} and {IIASA}},
  title = {Global Agro-Ecological Zones version 4 ({GAEZ} v4)},
  howpublished = {\url{http://www.fao.org/gaez/}},
  note = {Accessed: November 12, 2025},
  year = {2025}
}

@inproceedings{garnot2021panoptic,
  title={Panoptic segmentation of satellite image time series with convolutional temporal attention networks},
  author={Garnot, Vivien Sainte Fare and Landrieu, Loic},
  booktitle={Proceedings of the IEEE/CVF International Conference on Computer Vision},
  pages={4872--4881},
  year={2021}
}

@inproceedings{russwurm2019breizhcrops,
  title={{BreizhCrops}: A satellite time series dataset for crop type identification},
  author={Ru{\ss}wurm, Marc and Lef{\`e}vre, S{\'e}bastien and K{\"o}rner, Marco},
  booktitle={Proceedings of the International Conference on Machine Learning Time Series Workshop},
  volume={3},
  year={2019}
}

@misc{worldcerealsbenchmark,
  title = {Mapping crops at global scale! What works and what doesn't?},
  howpublished = {\url{https://blog.vito.be/remotesensing/worldcereal-benchmarking}},
  note = {Accessed: 2023-07-31},
  year={2021},
  author={Van Tricht, Kristof}
}

@article{helber2019eurosat,
  title={Eurosat: A novel dataset and deep learning benchmark for land use and land cover classification},
  author={Helber, Patrick and Bischke, Benjamin and Dengel, Andreas and Borth, Damian},
  journal={IEEE Journal of Selected Topics in Applied Earth Observations and Remote Sensing},
  volume={12},
  number={7},
  pages={2217--2226},
  year={2019},
  publisher={IEEE}
}

@inproceedings{lacoste2024geo,
  title={{GEO-Bench}: Toward foundation models for earth monitoring},
  author={Lacoste, Alexandre and Lehmann, Nils and Rodriguez, Pau and Sherwin, Evan and Kerner, Hannah and L{\"u}tjens, Bj{\"o}rn and Irvin, Jeremy and Dao, David and Alemohammad, Hamed and Drouin, Alexandre and others},
  booktitle={Advances in Neural Information Processing Systems},
  volume={36},
  year={2024}
}

@article{kikaki2024detecting,
  title={Detecting Marine pollutants and Sea Surface features with Deep learning in {Sentinel-2} imagery},
  author={Kikaki, Katerina and Kakogeorgiou, Ioannis and Hoteit, Ibrahim and Karantzalos, Konstantinos},
  journal={ISPRS Journal of Photogrammetry and Remote Sensing},
  volume={210},
  pages={39--54},
  year={2024},
  publisher={Elsevier}
}

@article{jaccard1901etude,
  title={{\'E}tude comparative de la distribution florale dans une portion des Alpes et des Jura},
  author={Jaccard, Paul},
  journal={Bull Soc Vaudoise Sci Nat},
  volume={37},
  pages={547--579},
  year={1901}
}

@article{shelhamer2016fully,
	author = {Shelhamer*, Evan and Long*, Jonathan and Darrell, Trevor},
	journal = {PAMI},
	title = {Fully Convolutional Networks for Semantic Segmentation},
	year = {2016}}

@article{ditommaso2021combining,
  title = {Combining GEDI and Sentinel-2 for wall-to-wall mapping of tall and short crops},
  author = {Di Tommaso, Stefania and Wang, Sherrie and Lobell, David B.},
  journal = {Environmental Research Letters},
  volume = {16},
  number = {12},
  pages = {125002},
  year = {2021},
}

@inproceedings{rao2025using,
  title={Using Multiple Input Modalities can Improve Data-Efficiency and OOD Generalization for ML with Satellite Imagery},
  author={Rao, Arjun and Rolf, Esther},
  booktitle={TerraBytes-Towards global datasets and models for Earth Observation},
  pages={166--188},
  year={2025},
  organization={PMLR}
}

@article{cong2022satmae,
  title={Satmae: Pre-training transformers for temporal and multi-spectral satellite imagery},
  author={Cong, Yezhen and Khanna, Samar and Meng, Chenlin and Liu, Patrick and Rozi, Erik and He, Yutong and Burke, Marshall and Lobell, David and Ermon, Stefano},
  journal={Advances in Neural Information Processing Systems},
  volume={35},
  pages={197--211},
  year={2022}
}

@inproceedings{
tseng2025galileo,
title={Galileo: Learning Global \& Local Features of Many Remote Sensing Modalities},
author={Gabriel Tseng and Anthony Fuller and Marlena Reil and Henry Herzog and Patrick Beukema and Favyen Bastani and James R Green and Evan Shelhamer and Hannah Kerner and David Rolnick},
booktitle={Forty-second International Conference on Machine Learning},
year={2025},
}

@inproceedings{jakubik2025terramind,
      title={{TerraMind}: Large-Scale Generative Multimodality for Earth Observation}, 
      author={Johannes Jakubik and Felix Yang and Benedikt Blumenstiel and Erik Scheurer and Rocco Sedona and Stefano Maurogiovanni and Jente Bosmans and Nikolaos Dionelis and Valerio Marsocci and Niklas Kopp and Rahul Ramachandran and Paolo Fraccaro and Thomas Brunschwiler and Gabriele Cavallaro and Juan Bernabe-Moreno and Nicolas Longépé},
      year={2025},
      booktitle={ICCV},
      eprint={2504.11171},
      archivePrefix={arXiv},
      primaryClass={cs.CV},
      url={https://arxiv.org/abs/2504.11171}, 
}

@article{herzog2025olmoearth,
  title={OlmoEarth: Stable Latent Image Modeling for Multimodal Earth Observation},
  author={Herzog, Henry and Bastani, Favyen and Zhang, Yawen and Tseng, Gabriel and Redmon, Joseph and Sablon, Hadrien and Park, Ryan and Morrison, Jacob and Buraczynski, Alexandra and Farley, Karen and others},
  journal={arXiv preprint arXiv:2511.13655},
  year={2025}
}

@inproceedings{waldmann2025panopticon,
  title={Panopticon: Advancing any-sensor foundation models for earth observation},
  author={Waldmann, Leonard and Shah, Ando and Wang, Yi and Lehmann, Nils and Stewart, Adam and Xiong, Zhitong and Zhu, Xiao Xiang and Bauer, Stefan and Chuang, John},
  booktitle={Proceedings of the Computer Vision and Pattern Recognition Conference},
  pages={2204--2214},
  year={2025}
}

@inproceedings{astruc2025anysat,
  title={AnySat: One Earth Observation Model for Many Resolutions, Scales, and Modalities},
  author={Astruc, Guillaume and Gonthier, Nicolas and Mallet, Clement and Landrieu, Loic},
  booktitle={Proceedings of the Computer Vision and Pattern Recognition Conference},
  pages={19530--19540},
  year={2025}
}

@inproceedings{mindermann2022prioritized,
  title={Prioritized training on points that are learnable, worth learning, and not yet learnt},
  author={Mindermann, S{\"o}ren and Brauner, Jan M and Razzak, Muhammed T and Sharma, Mrinank and Kirsch, Andreas and Xu, Winnie and H{\"o}ltgen, Benedikt and Gomez, Aidan N and Morisot, Adrien and Farquhar, Sebastian and others},
  booktitle={International Conference on Machine Learning},
  pages={15630--15649},
  year={2022},
  organization={PMLR}
}

@inproceedings{bastani2023satlaspretrain,
  title={{SatlasPretrain}: A large-scale dataset for remote sensing image understanding},
  author={Bastani, Favyen and Wolters, Piper and Gupta, Ritwik and Ferdinando, Joe and Kembhavi, Aniruddha},
  booktitle={Proceedings of the IEEE/CVF International Conference on Computer Vision},
  pages={16772--16782},
  year={2023}
}

@misc{planet2022planet,
  author =    {Planet Labs PBC},
  organization = {Planet},
  title =     {Planet Application Program Interface: In Space for Life on Earth},
  year =      {2020--2022},
  url = "https://api.planet.com"
}

@article{houriez2025scalable,
  title={Scalable geospatial data generation using AlphaEarth foundations model},
  author={Houriez, Luc and Pilarski, Sebastian and Vahedi, Behzad and Ahmadalipour, Ali and Scully, Teo Honda and Aflitto, Nicholas and Andre, David and Jaffe, Caroline and Wedner, Martha and Mazzola, Rich and others},
  journal={arXiv preprint arXiv:2508.11739},
  year={2025}
}

@article{ma2025harvesting,
  title={Harvesting AlphaEarth: Benchmarking the Geospatial Foundation Model for Agricultural Downstream Tasks},
  author={Ma, Yuchi and Shen, Yawen and Swatantran, Anu and Lobell, David B},
  journal={arXiv preprint arXiv:2601.00857},
  year={2025}
}

@article{de2023haie,
  title={La haie, levier de la planification {\'e}cologique},
  author={de Menthi{\`e}re, Catherine and Falcone, Patrick and Piveteau, Vincent and Ory, Xavier},
  journal={Rapport du CGAAER},
  volume={22114},
  pages={116},
  year={2023}
}

@article{baudry2000hedgerows,
  title={Hedgerows: an international perspective on their origin, function and management},
  author={Baudry, Jacques and Bunce, RGH and Burel, Fran{\c{c}}oise},
  journal={Journal of environmental management},
  volume={60},
  number={1},
  pages={7--22},
  year={2000},
  publisher={Elsevier}
}

@article{cantelaube2014registre,
  title={Le registre parcellaire graphique: des donn{\'e}es g{\'e}ographiques pour d{\'e}crire la couverture du sol agricole},
  author={Cantelaube, Pierre and Carles, Marie},
  journal={NOV'AE-Ing{\'e}nierie et savoir-faire innovants},
  number={sp{\'e}cial Cahier des techniques},
  pages={58--64},
  year={2014}
}

@techreport{IGN_OFB_2020_BDHaie,
  author      = {IGN and OFB},
  title       = {Les Haies du Dispositif de Suivi des Bocages},
  institution = {Institut national de l'information géographique 
                 et forestière (IGN) / Office français de la 
                 biodiversité (OFB)},
  year        = {2020},
  month       = {12},
  type        = {Rapport technique},
  url         = {https://geoservices.ign.fr/sites/default/files/
                 2021-07/Descriptif_de_contenu_et_limite_DSB.pdf},
  urldate     = {2026-03-26},
  note        = {Version 1.0}
}

@techreport{Commagnac_2024_BDHaieV2,
  author      = {Commagnac, Lo{\"i}c and Letouze, Fr{\'e}d{\'e}ric 
                 and {Le Bellec}, Louise},
  title       = {Dispositif National de Suivi des Bocages : 
                 mise à jour du référentiel des haies établi 
                 lors de la phase 1},
  institution = {Institut national de l'information géographique 
                 et forestière (IGN) / Office français de la 
                 biodiversité (OFB)},
  year        = {2024},
  month       = {1},
  type        = {Rapport technique},
  url         = {https://geoservices.ign.fr/sites/default/files/
                 2025-06/Descriptif_de_contenu_et_limite_DSB_v2.pdf},
  urldate     = {2026-03-26},
  note        = {Version 1.0, Conventions MTECT--OFB--IGN}
}
